\pdfoutput=1
\documentclass[11pt]{article}

\usepackage[preprint]{acl}

\usepackage{times}
\usepackage{latexsym}
\usepackage[T1]{fontenc}
\usepackage[utf8]{inputenc}
\usepackage{microtype}
\usepackage{inconsolata}
\usepackage{graphicx}
\usepackage{booktabs}
\usepackage{adjustbox}
\usepackage{siunitx}
\usepackage{url}
\usepackage{multirow}
\usepackage{tcolorbox}
\usepackage{amsmath}

\definecolor{promptbackground}{RGB}{198 207 215}
\definecolor{promptborder}{RGB}{120 163 199}

\newtcolorbox{painit}[1][Problem Analyst Initialisation Prompt]{%
  title={\textbf{#1}},
  colback=promptbackground,
  colframe=promptborder,
  sharp corners,
  boxrule=0.5pt,
  top=5pt,
  bottom=5pt,
  left=5pt,
  right=5pt,
  width=\textwidth, 
  enlarge left by=0mm, 
  enlarge right by=0mm, 
  before upper={\noindent} 
}
\newtcolorbox{sainit}[1][Solution Analyst Initialisation Prompt]{%
  title={\textbf{#1}},
  colback=promptbackground,
  colframe=promptborder,
  sharp corners,
  boxrule=0.5pt,
  top=5pt,
  bottom=5pt,
  left=5pt,
  right=5pt,
  width=\textwidth, 
  enlarge left by=0mm, 
  enlarge right by=0mm, 
  before upper={\noindent} 
}

\newtcolorbox{cainit}[1][Criterion Analyst Initialisation Prompt]{%
  title={\textbf{#1}},
  colback=promptbackground,
  colframe=promptborder,
  sharp corners,
  boxrule=0.5pt,
  top=5pt,
  bottom=5pt,
  left=5pt,
  right=5pt,
  width=\textwidth, 
  enlarge left by=0mm, 
  enlarge right by=0mm, 
  before upper={\noindent} 
}

\newtcolorbox{pad}[1][Problem Analyst Discussion Prompt]{%
  title={\textbf{#1}},
  colback=promptbackground,
  colframe=promptborder,
  sharp corners,
  boxrule=0.5pt,
  top=5pt,
  bottom=5pt,
  left=5pt,
  right=5pt,
  width=\textwidth, 
  enlarge left by=0mm, 
  enlarge right by=0mm, 
  before upper={\noindent} 
}

\newtcolorbox{sad}[1][Solution Analyst Discussion Prompt]{%
  title={\textbf{#1}},
  colback=promptbackground,
  colframe=promptborder,
  sharp corners,
  boxrule=0.5pt,
  top=5pt,
  bottom=5pt,
  left=5pt,
  right=5pt,
  width=\textwidth, 
  enlarge left by=0mm, 
  enlarge right by=0mm, 
  before upper={\noindent} 
}

\newtcolorbox{nov}[1][Instructions for human annotators for novelty ground truth]{%
  title={\textbf{#1}},
  colback=promptbackground,
  colframe=promptborder,
  sharp corners,
  boxrule=0.5pt,
  top=5pt,
  bottom=5pt,
  left=5pt,
  right=5pt,
  width=\textwidth, 
  enlarge left by=0mm, 
  enlarge right by=0mm, 
  before upper={\noindent} 
}

\newtcolorbox{cad}[1][Criterion Analyst Discussion Prompt]{%
  title={\textbf{#1}},
  colback=promptbackground,
  colframe=promptborder,
  sharp corners,
  boxrule=0.5pt,
  top=5pt,
  bottom=5pt,
  left=5pt,
  right=5pt,
  width=\textwidth, 
  enlarge left by=0mm, 
  enlarge right by=0mm, 
  before upper={\noindent} 
}

\newtcolorbox{instructions}[1][Instructions for human annotators for convergent creativity ground truth]{%
  title={\textbf{#1}},
  colback=promptbackground,
  colframe=promptborder,
  sharp corners,
  boxrule=0.5pt,
  top=5pt,
  bottom=5pt,
  left=5pt,
  right=5pt,
  width=\textwidth, 
  enlarge left by=0mm, 
  enlarge right by=0mm, 
  before upper={\noindent} 
}

\newtcolorbox{novelty}[1][Instructions for human annotators for novelty ground truth]{%
  title={\textbf{#1}},
  colback=promptbackground,
  colframe=promptborder,
  sharp corners,
  boxrule=0.5pt,
  top=5pt,
  bottom=5pt,
  left=5pt,
  right=5pt,
  width=\textwidth, 
  enlarge left by=0mm, 
  enlarge right by=0mm, 
  before upper={\noindent} 
}

\newtcolorbox{conf}[1][Confidence Prompt]{%
  title={\textbf{#1}},
  colback=promptbackground,
  colframe=promptborder,
  sharp corners,
  boxrule=0.5pt,
  top=5pt,
  bottom=5pt,
  left=5pt,
  right=5pt
}

\newtcolorbox{verdict}[1][Verdict Prompt]{%
  title={\textbf{#1}},
  colback=promptbackground,
  colframe=promptborder,
  sharp corners,
  boxrule=0.5pt,
  top=5pt,
  bottom=5pt,
  left=5pt,
  right=5pt
}

\newtcolorbox{oneshot}[1][Oneshot Prompt]{%
  title={\textbf{#1}},
  colback=promptbackground,
  colframe=promptborder,
  sharp corners,
  boxrule=0.5pt,
  top=5pt,
  bottom=5pt,
  left=5pt,
  right=5pt
}

\newtcolorbox{fewshotcot}[1][Fewshot + Chain-of-Thought Prompt]{%
  title={\textbf{#1}},
  colback=promptbackground,
  colframe=promptborder,
  sharp corners,
  boxrule=0.5pt,
  top=5pt,
  bottom=5pt,
  left=5pt,
  right=5pt
}

\newtcolorbox{fewshot}[1][Fewshot Prompt]{%
  title={\textbf{#1}},
  colback=promptbackground,
  colframe=promptborder,
  sharp corners,
  boxrule=0.5pt,
  top=5pt,
  bottom=5pt,
  left=5pt,
  right=5pt
}

\newtcolorbox{cott}[1][Chain-of-Thought Prompt]{%
  title={\textbf{#1}},
  colback=promptbackground,
  colframe=promptborder,
  sharp corners,
  boxrule=0.5pt,
  top=5pt,
  bottom=5pt,
  left=5pt,
  right=5pt
}

\newtcolorbox{mad}[1][Multi-agent Debate - Debater Prompt]{%
  title={\textbf{#1}},
  colback=promptbackground,
  colframe=promptborder,
  sharp corners,
  boxrule=0.5pt,
  top=5pt,
  bottom=5pt,
  left=5pt,
  right=5pt
}

\newtcolorbox{madjuj}[1][Multi-agent Debate - Judge Prompt]{%
  title={\textbf{#1}},
  colback=promptbackground,
  colframe=promptborder,
  sharp corners,
  boxrule=0.5pt,
  top=5pt,
  bottom=5pt,
  left=5pt,
  right=5pt
}

\newtcolorbox{noveltyjuj}[1][Novelty Judge Prompts]{%
  title={\textbf{#1}},
  colback=promptbackground,
  colframe=promptborder,
  sharp corners,
  boxrule=0.5pt,
  top=5pt,
  bottom=5pt,
  left=5pt,
  right=5pt,
  width=\textwidth, 
  enlarge left by=0mm, 
  enlarge right by=0mm, 
  before upper={\noindent} 
}

\newtcolorbox{backgroundhighlight}[1][]{
  colback=green!10!white,
  boxrule=0.5pt,           
  arc=0mm,                 
  breakable,
  #1
}
\title{Automated Creativity Evaluation of Language Models \\Across Open-Ended Tasks}

\author{
  \textbf{Tan Min Sen\textsuperscript{1,*,\dag}},
  \textbf{Zachary Choy Kit Chun\textsuperscript{1,*}},
  \textbf{Syed Ali Redha Alsagoff\textsuperscript{2}}, 
  \\
  \textbf{Nadya Yuki Wangsajaya\textsuperscript{2}},
  \textbf{Banerjee Mohor\textsuperscript{2}},
  \textbf{Swaagat Bikash Saikia\textsuperscript{1}},
  \textbf{Alvin Chan\textsuperscript{2,3,4,\dag}}
  \\[0.5em]
\textsuperscript{1}Raffles Institution\\
\textsuperscript{2}College of Computing and Data Science, Nanyang Technological University\\
\textsuperscript{3}Lee Kong Chian School of Medicine, Nanyang Technological University\\
\textsuperscript{4}Centre of AI in Medicine (C-AIM), Nanyang Technological University
  \\[0.3em]
  \textsuperscript{*}\small Equal contribution \quad
  \textsuperscript{\dag}\small Corresponding authors: 
  \href{mailto:guoweialvin.chan@ntu.edu.sg}{guoweialvin.chan@ntu.edu.sg},
  \href{mailto:minsen@tanminsen.com}{minsen@tanminsen.com}
}

\begin{document}
\maketitle

\begin{abstract}
Large language models (LLMs) have achieved remarkable progress in language understanding, reasoning, and generation, sparking growing interest in their creative potential. Realizing this potential requires systematic and scalable methods for evaluating creativity across diverse tasks. However, most existing creativity metrics are tightly coupled to specific tasks, embedding domain assumptions into the evaluation process, and limiting scalability and generality. To address this gap, we introduce an automated, domain-agnostic framework for quantifying LLM creativity across open-ended tasks. Our approach separates the measurement apparatus from the creative task itself, enabling scalable, task-agnostic assessment. Divergent creativity is measured using semantic entropy, a reference-free and robust metric for novelty and diversity, validated against human annotations, LLM-based novelty judgments and baseline diversity measures. Convergent creativity is assessed via a novel retrieval-based multi-agent judge framework that delivers context-sensitive evaluation of task fulfilment with over 60\% improved efficiency. We validate our framework in three qualitatively distinct domains: problem-solving (MacGyver), research ideation (HypoGen), and creative writing (BookMIA), using a broad suite of LLMs. Empirical results show that our framework reliably captures key facets of creativity, including novelty, diversity, and task fulfilment, and reveal how model properties, such as size, temperature, recency, and reasoning, impact creative performance. Our work establishes a reproducible and generalizable standard for automated LLM creativity evaluation, paving the way for scalable benchmarking and accelerating progress in creative AI.\footnote{Our code is available at \url{https://github.com/tanminsen/creativity-eval}.}
\end{abstract}

\newpage
\section{Introduction}

\paragraph{} Recent advances in large language models (LLMs) have led to major breakthroughs in language comprehension, generation, and reasoning \citep{lewis2019EA, manning2022}. As LLMs become more adept at reasoning, their creative potential has emerged as a key area of interest \citep{ye2024EA, sun2024EA}. Creative LLMs can accelerate scientific discovery by proposing unconventional solutions \citep{ruan2024EA, gu2024llmsrealizecombinatorialcreativity}, uncovering novel patterns \citep{si2024llmsgeneratenovelresearch}, and automating experiment design \citep{liu2024fullyautonomousresearchpowered}. Understanding and quantifying these creative capabilities is thus increasingly important.

However, most existing creativity evaluation frameworks are tightly coupled to specific tasks or domains, embedding strong domain assumptions into the assessment process \citep{ye2024EA, delorenzo2024EA}. These approaches often rely on curated answer sets, hand-crafted rubrics, or extensive human annotation, rendering creativity assessment subjective, resource-intensive, and difficult to scale. As a result, the field lacks automated, domain-general evaluation standards.

To address these challenges, we propose a fully automated, domain-general framework for evaluating LLM creativity that is both robust and scalable across open-ended tasks. Our framework decouples evaluation from specific creative tasks, enabling systematic, reference-free assessment of model creativity across domains. Building on cognitive science, which characterizes creativity as encompassing both divergent and convergent thinking \citep{guilford1950}, we deliberately design our framework to evaluate both aspects through novel, automated methods. This separation distinguishes creative breadth from task fulfilment in open-ended settings. 

Divergent thinking is the ability to generate diverse, novel, and innovative ideas. We hypothesize that divergent creativity is reflected in the diversity of plausible semantic directions a model can explore in open-ended tasks. To operationalize this hypothesis, we adapt \emph{semantic entropy}, a sampling-based, reference-free metric that quantifies variability in model-generated outputs. We validate its utility by benchmarking semantic entropy against human annotations, LLM-based novelty judgments, and other diversity baselines, finding that it faithfully reflects core markers of divergent creativity.

Convergent thinking involves synthesizing information to produce solutions tailored to specific goals and contexts \citep{kumar2024EA}. Recognizing the inherent subjectivity in evaluating this aspect \citep{li2023EA}, we propose an adaptable, autonomous multi-agent LLM judging framework, where agents collaboratively assess distinct facets of task fulfilment \citep{lu2024EAllmdiscussionenhancingcreativity}. To address the computational inefficiency of traditional discussion-based evaluations \citep{wang2024EAbudgetefficiency}, we introduce a retrieval-based discussion framework that streamlines the review process, making large-scale benchmarking more feasible.

To demonstrate the generality and practical value of our framework, we evaluate it across three qualitatively distinct domains of creativity: \textbf{problem-solving} with the MacGyver dataset \citep{macgyver}, \textbf{research ideation} with the HypoGen dataset \citep{oneill2025sparkssciencehypothesisgeneration}, and \textbf{artistic creativity} through creative writing with the BookMIA dataset \citep{shi2024detectingpretrainingdatalarge}. Together, these domains capture complementary facets of creativity—functional reasoning under constraints, logical synthesis of scientific ideas, and narrative originality. We apply both methods to 300 problems per domain and benchmark diverse LLMs, analyzing how model size, recency, temperature, and reasoning augmentation affect creativity.

In summary, we: \textbf{(1)} introduce a reference-free, automated assessment of divergent creativity via semantic entropy; \textbf{(2)} develop a more compute-efficient multi-agent LLM judging framework for evaluating convergent creativity; and \textbf{(3)} present comprehensive empirical benchmarking across MacGyver, HypoGen, and BookMIA, together establishing an automated, domain-general framework for LLM creativity evaluation. Our experiments demonstrate that the proposed framework reliably captures complementary facets of creativity, generalizes across diverse domains, and enables systematic analysis of how model size, recency, temperature, and reasoning augmentation influence creative performance.

\section{Background}\label{sec:Background}

\paragraph{Human Creativity Tests.} Classic human creativity assessments, such as Torrance Tests of Creative Thinking (TTCT) and Consensual Assessment Technique (CAT) \citep{torranceND, cat}, have been adapted to evaluate LLMs. However, these methods depend on extensive human annotation, making them unscalable and ill-suited for automated evaluation. Moreover, while TTCT metrics like fluency and elaboration are meaningful in human settings, they are less reliable for LLMs, since idea count and output length can be trivially adjusted by sampling. Consequently, our framework focuses on the TTCT dimensions of originality and flexibility as measures of divergent creativity, while using a separate automated judge to assess task fulfilment through convergent creativity.

\paragraph{Domain-specific Creativity Evaluation.} Beyond classic human tests, a wide range of task-specific creativity benchmarks have been developed for LLMs, like mathematical reasoning, hardware design, metaphor generation and coding \citep{ye2024EA, delorenzo2024EA, distefano2024EA, gomez-rodriguezandwilliams2023}. These frameworks typically embed strong domain assumptions, require curated answer sets or subjective metrics, and are closely tied to the structure of their target tasks. Hence, they lack generalizability and are difficult to apply systematically to the open-ended challenges tackled by LLMs. Our approach overcomes these limitations by providing a task-agnostic, reference-free, and fully automated method for creativity evaluation.

\paragraph{Divergent Creativity Evaluation.} Automated metrics for divergent creativity in LLMs, including semantic similarity, integration scores, and Lempel–Ziv complexity, offer some insight into output diversity, but often miss the nuance required for complex, open-ended tasks \citep{mohammadi2024, chenandding2023, summersstay2023EA, peeperkorn2024EA, bellemarepepin2024EA}. Recent work has instead used uncertainty to detect hallucinations in LLM outputs \citep{huang2024EA, chen2025enhancinguncertaintymodelingsemantic, zhang2023EA, sriramanan2024EA}, including one based on Semantic Entropy (SE)\citep{SE}. We build on this line of work by reinterpreting Semantic Entropy, originally proposed for hallucination detection, as a reference-free operationalization of divergent creativity in open-ended generation.

\begin{figure*}[t]
  \centering
  \vspace{-2cm}
  \includegraphics[width=\textwidth]{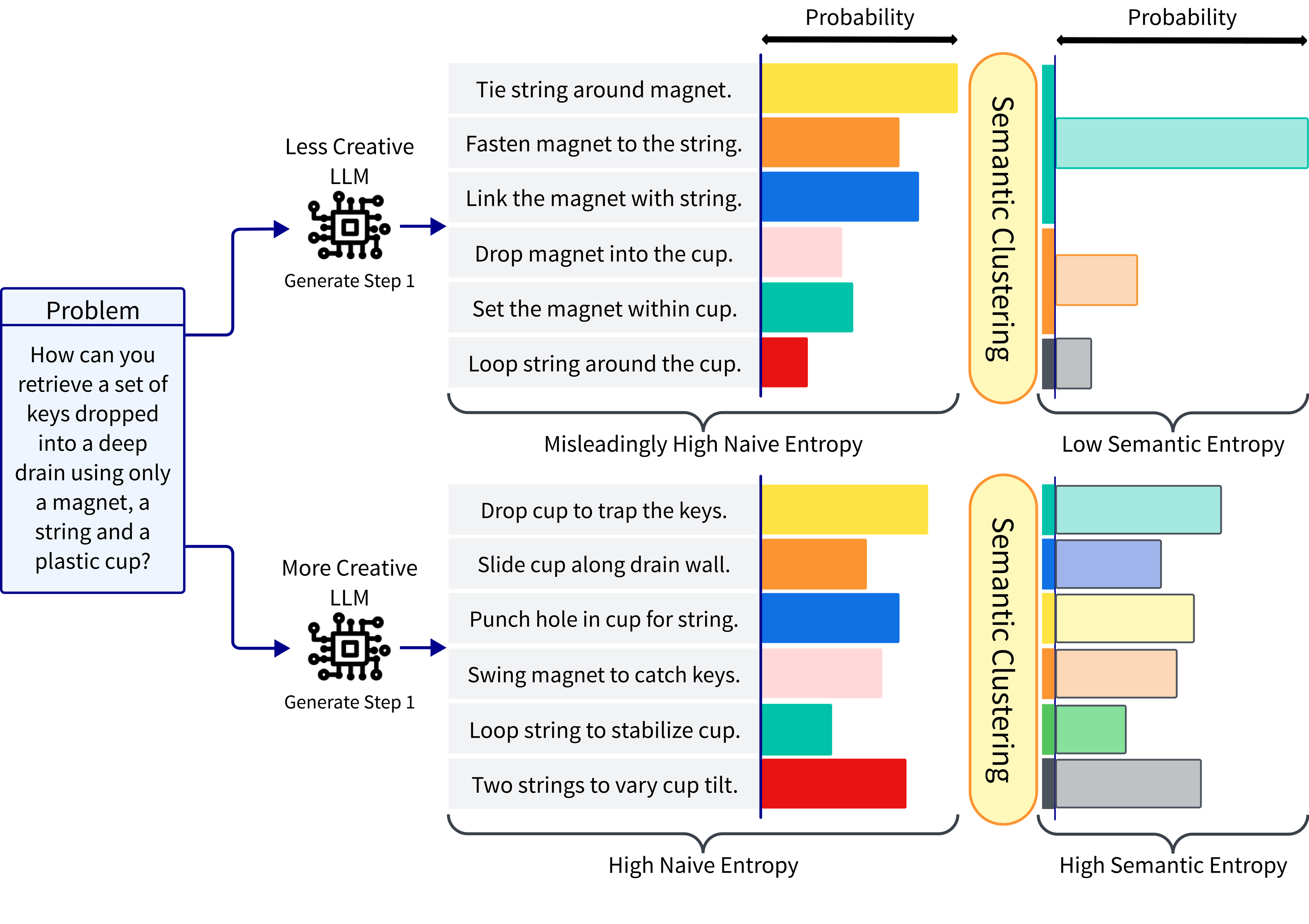}
  \vspace{-1.3cm}
  \caption{Divergent Creativity. LLM-generated steps clustered by similarity, with entropy computed over cluster probabilities. Naive entropy (middle) uses raw probabilities, while semantic entropy (right) clusters by meaning. \textbf{Top:} a less creative model yields surface-level rephrasings and low semantic entropy. \textbf{Bottom:} a more creative model yields genuinely distinct ideas and high semantic entropy. }
  \label{fig:sediag}
  \vspace{-0.3cm}
\end{figure*}

\newpage
\paragraph{Convergent Creativity Evaluation.} Convergent creativity is the ability to refine, select, and deliver solutions that are  useful, feasible, and context-appropriate. Foundational creativity research \citep{torranceND, guilford1950, Cropley01072006, amabile, Runco01012012} consistently defines creativity as producing ideas that are both novel (divergent) and useful or appropriate (convergent). While divergent creativity captures the generation of original and varied ideas, convergent creativity evaluates whether those ideas effectively solve the problem, satisfy constraints, and are practically viable. This distinction is critical, since models may produce diverse outputs that are incoherent or irrelevant; without convergent evaluation, such outputs would misleadingly be perceived as “creative.”

Traditional tests like the Remote Associates Test \citep{Mednick1967RemoteAT} are ill-suited to LLMs, as they were designed for humans. Recent pipelines leverage LLMs as judges \citep{liu-etal-2023-g, zhu2025judgelmfinetunedlargelanguage, kim2024prometheusinducingfinegrainedevaluation}, with multi-agent discussion frameworks shown to provide more nuanced and comprehensive evaluation of model outputs \citep{liang2024EA, chan2023EA}. However, these methods are computationally intensive and hard to scale \citep{lv2024EA, luo2023EA}. Our retrieval-based discussion framework addresses this, enabling scalable, robust evaluation of task fulfilment. 

\section{Divergent Creativity}
\subsection{Motivation and Conceptual Framing of Semantic Entropy}

Creative reasoning tasks are often open-ended and admit multiple plausible, simultaneously valid solutions, rather than a single correct answer. In such settings, divergent creativity is characterized by the breadth and novelty of ideas a model can generate, reflected in its ability to explore distinct solution directions from the same problem context.

To quantify this property in model outputs, we adapt \textbf{Semantic Entropy (SE)} \citep{SE}. In the original work, SE was demonstrated primarily in \textbf{single–gold-answer} question answering settings, where uncertainty reflects a model’s lack of confidence in the expected correct response and was therefore introduced as a calibrated indicator of hallucination. 

In contrast, the creative reasoning tasks considered here are inherently open-ended and do not admit a single gold answer. Under this regime, probability mass distributed across semantically distinct ideas need not indicate error; instead, it may reflect a model’s exploration of alternative solution paths. We therefore hypothesize that, in open-ended settings, SE captures the breadth of a model’s generative exploration by measuring how probability mass is allocated across semantic classes (Fig. \ref{fig:sediag}).

\begin{figure*}[t]
\centering
\vspace{-2cm}
\includegraphics[width=\textwidth]{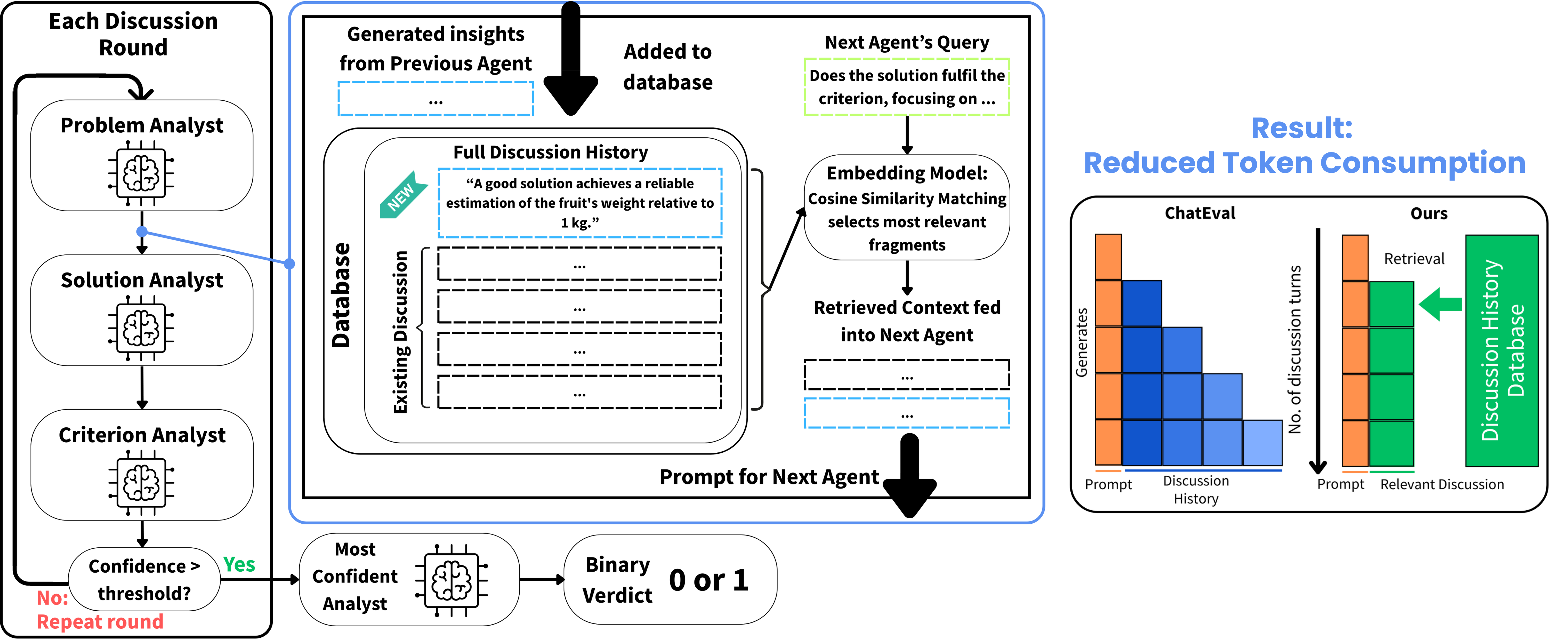} 
\vspace{-0.75cm}
\caption{Retrieval-Based Multi-Agent Framework. \textbf{Left:} Three specialised LLM agents---Problem, Solution, and Criterion---analyze tasks from different perspectives, recording insights in "fragments". \textbf{Middle:} Fragments are embedded in a vector database; each agent retrieves the $k$ most relevant fragments via cosine similarity at their turn. \textbf{Right:} This retrieval loop cuts token usage by $\approx$63\% compared to ChatEval while converging on a binary verdict.}
\label{fig:discussionframework}
\vspace{-0.35cm}
\end{figure*}

\subsection{Semantic Entropy Formulation}

\paragraph{Semantic Clustering.} Following \citet{SE}, Step generations \(({s_1...s_n})\)  are clustered using bi-directional entailment, where a greedy algorithm assigns each generation to an existing class \(C_a\) if similar, or creates a new class otherwise. 

\paragraph{Semantic Entropy.} For a query \(x\), the probability, \(P(s|x)\), of a generated step\(s\), comprising tokens \((t_{1},...,t_{i})\) is given by the product of its conditional token probabilities. For computational efficiency, we use log-probability \(\log P(s|x)\).  

\begin{equation}
    \log P(s|x) = \sum_{i}^{}\log P(t_{i}|t_{< i},x)\label{eqn1} 
\end{equation}

The probability of a semantic class \(c\) is the sum of all generated samples \(s\) belonging to the class: 

\begin{equation}
    P(c|x)\ = \ \sum_{s \in c}^{}P(s|x)\label{eqn2}
\end{equation}

Semantic Entropy is computed as the entropy of the class probability distribution over all classes \(C\):

\vspace{-0.3cm}
\begin{equation}
    H(x) = - \sum_{i = 1}^{|C|}P(C_{i}|x)\log P(C_{i}|x)\label{eqn3}
\end{equation}

\paragraph{Implementation.} We compute SE in a stepwise manner. At each generation step, we sample \(n = 10\) candidate solutions, cluster them by semantic equivalence, and compute SE over the resulting class probability distribution (Fig.~\ref{fig:sediag}). Crucially, SE operates at the level of meaning rather than surface form, capturing genuine conceptual differences and avoiding inflation from superficial paraphrasing that affects word-level diversity metrics. Semantic clustering details, including entailment model selection and validation against human-annotated ground truth, are provided in Appendix~\ref{entailmentdetails}.

\section{Convergent Creativity}

\paragraph{Motivation and Challenges.} Evaluating convergent creativity requires assessing whether a candidate solution fulfils task requirements across multiple dimensions, such as feasibility, coherence, relevance, and domain-specific correctness. As discussed in Section \ref{sec:Background}, distributing evaluative responsibilities across multiple specialized LLM agents yields more comprehensive and human-aligned judgments than single-judge evaluation \citep{chan2023EA, liang2024EA}. However, existing multi-agent frameworks, most notably ChatEval, append the full discussion history at each turn, causing prompt length and token cost to grow with each round of deliberation. This makes them prohibitively expensive to deploy at benchmark scale, particularly for multi-step, open-ended tasks, highlighting that scalability, rather than evaluation quality, is the primary bottleneck for multi-agent convergent creativity assessment.

\paragraph{Scalable Convergent Creativity Evaluation.}To address this limitation, we introduce a retrieval-based multi-agent judging framework (Fig.~\ref{fig:discussionframework}) that maintains multi-agent deliberation while substantially reducing computational cost. Instead of replaying the full discussion history, agents store intermediate analyses as retrievable fragments and attend only to the most relevant prior discussion fragments at each turn. This design prevents unbounded context growth while maintaining agent specialization, reducing token usage by 63\% vs. ChatEval (see Appendix~\ref{llmjcosts}) without sacrificing accuracy. As a result, our framework enables practical, large-scale evaluation of convergent creativity across open-ended benchmarks. Full prompt templates, mathematical formalization, and implementation details are provided in Appendix~\ref{sec:implementdetails}.

\begin{figure*}[t]
    \centering
    \vspace{-1.9cm}
        \includegraphics[width=\textwidth]{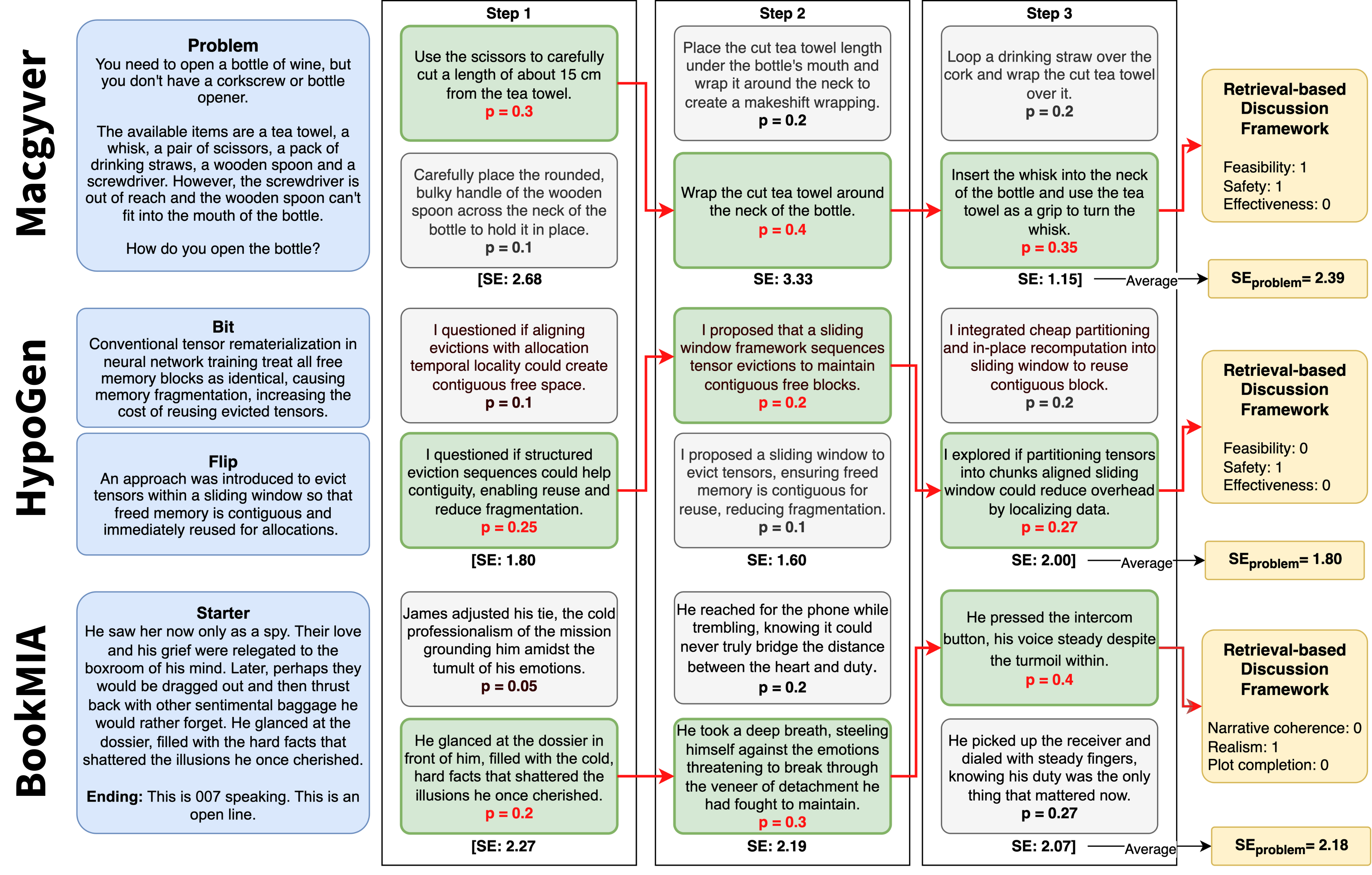}
        \vspace{-0.8cm}
        \caption{Overview of our benchmark, for all 3 datasets.}
    \vspace{-0.35cm}
    \label{fig:pipeline}
\end{figure*}

\newpage
\section{Experimental Setup}\label{sec:expt setup}

\paragraph{Domains and Evaluation Metrics.}
We evaluate our framework across three complementary creativity domains---unconventional problem-solving, research ideation, and artistic creativity---motivated by the taxonomy of \citet{ismayilzada2025creativityaiprogresseschallenges}: \textbf{MacGyver} \citep{macgyver}, \textbf{HypoGen} \citep{oneill2025sparkssciencehypothesisgeneration}, and \textbf{BookMIA} \citep{shi2024detectingpretrainingdatalarge}. These domains probe distinct facets of creativity: practical reasoning under constraints, synthesis of novel scientific hypotheses, and narrative originality.

\begin{itemize}

    \item \textbf{MacGyver:} Physical reasoning tasks requiring unconventional use of everyday objects.\\
    \emph{Feasibility, Safety, Effectiveness}

    \item \textbf{HypoGen:} Open-ended scientific ideation tasks, where models generate a novel reasoning chain from a given ``Bit'' to a target ``Flip'' without access to the ground-truth.\\
    \emph{Feasibility, Scientific Accuracy, Relevance}

    \item \textbf{BookMIA:} Creative writing tasks in which models generate multi-paragraph stories to connect provided starter and ending sentences.\\
    \emph{Coherence, Realism, Plot Completion}

\end{itemize}
\vspace{-0.1cm}
\noindent\emph{Prompts and metric definitions in Appendix~\ref{llmjmetrics}.}

\paragraph{Benchmarking Protocol.}We benchmark LLMs on the three datasets using a unified, step-wise evaluation pipeline (Fig.~\ref{fig:pipeline}). For each model, 300 problems per domain are solved incrementally: at each step, the model samples $n=10$ candidate continuations, which are clustered to compute SE. The highest-probability continuation is then selected via greedy decoding and appended to the solution, repeating until completion. Divergent creativity is reported as the average SE across all steps, while convergent creativity is assessed on the final solutions using our multi-agent judge and domain-specific metrics. Final model scores report both divergent (SE) and convergent (judge score).

\paragraph{Design Rationale.} This two-stage design, sampling multiple next-step continuations to measure divergent creativity and then applying greedy decoding to construct the highest-probability solution path, is intentional. 

Divergent creativity is estimated from the distribution of possible next-step continuations a model considers at each reasoning step, reflecting the range of approaches it explores rather than the creativity of a single final trajectory. This intuition aligns with existing LLM reasoning methods such as Tree-of-Thoughts \citep{yao2023treethoughtsdeliberateproblem}, where local branching behavior provides a practical approximation of a model’s exploration space. Greedy decoding is used for the convergent component, as convergent creativity concerns selecting and refining ideas to arrive at one effective or correct solution, and is therefore best evaluated through the model’s most confident answer.

\paragraph{} All experiments used 4 A100 GPUs, with API access for large models. See the Appendix for further details on model selection (\ref{appendixsth}), and controlled experiments evaluating choices on Semantic Entropy aggregation (\ref{SEaggregation}) and greedy decoding (\ref{sec:se_human_validation}).

\begin{figure*}[t]
\vspace{-1.9cm}
\centering
\includegraphics[width=\textwidth]{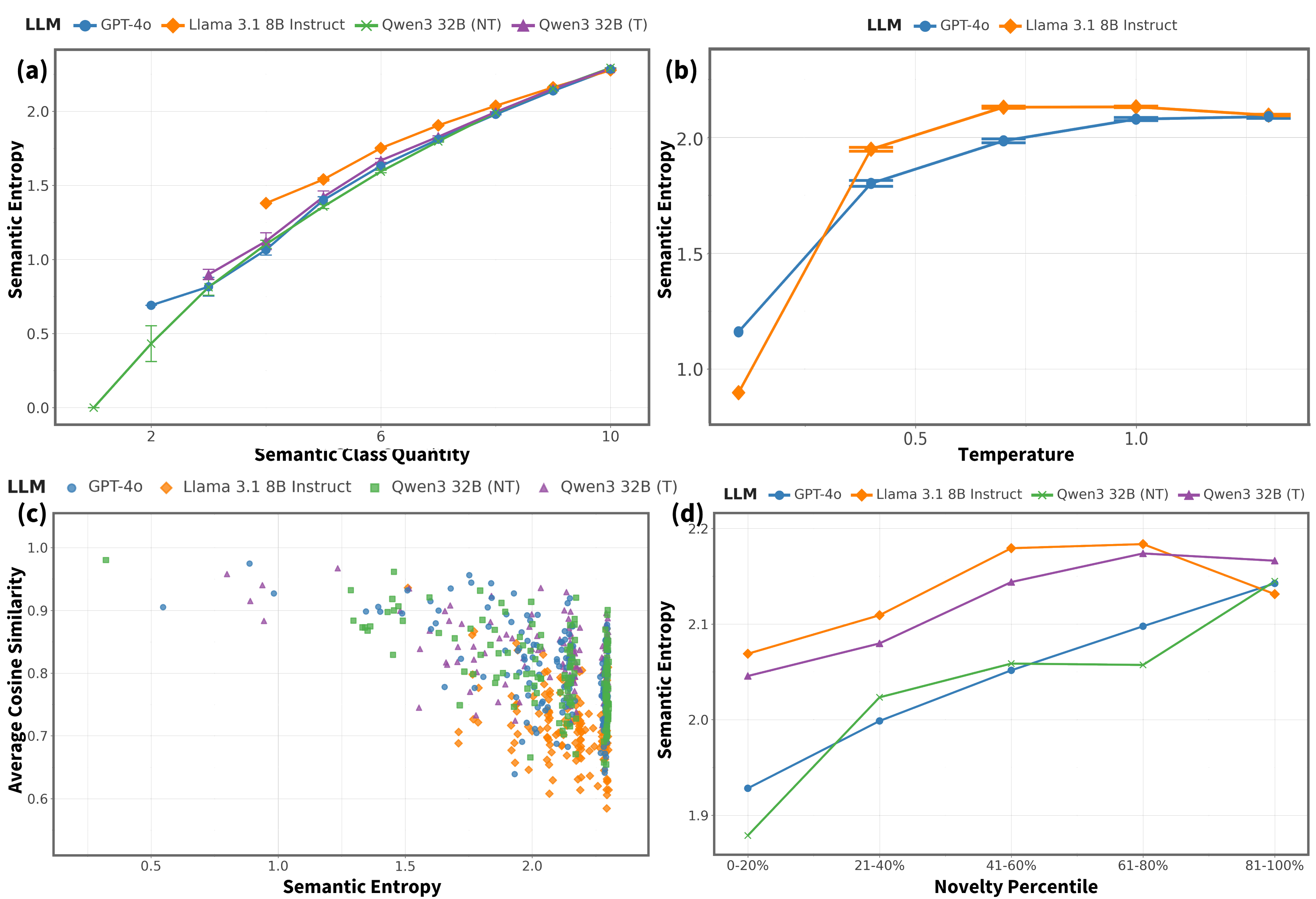}
\vspace{-0.7cm}
\caption{Semantic Entropy analysis. \textbf{(a):} Step-level SE against no. of semantic clusters formed at each step. \textbf{(b):} Average step-level SE under different sampling temperatures. \textbf{(c):} Step-level SE vs. average pairwise cosine similarity among sampled candidates. \textbf{(d):} Solution-level SE versus LLM-judged novelty rankings of final solutions. (T) and (NT) denote Thinking and Non-Thinking model variants, respectively.}
\label{fig:divergent}
\vspace{-0.3cm}
\end{figure*}

\section{Results and Discussion}\label{sec:results}

\subsection{Divergent Creativity }

\paragraph{Step-level Semantic Entropy behaves as a principled measure of local divergent exploration.} We find that step-level SE is strongly correlated with the number of unique idea categories a model considers at each generation step (Fig. 4a), indicating that it reliably reflects the creativity dimension of flexibility, one of the TTCT metrics. Step-level SE also increases with sampling temperature across models (Fig. 4b), consistent with findings that increased temperature leads to greater exploratory behavior \citep{chenandding2023}. Finally, step-level SE is negatively correlated with average pairwise cosine similarity among sampled candidates (Fig. 4c). Models with higher SE tend to produce sets of candidates that are less similar under embedding-based measures, indicating greater semantic diversity beyond surface-form variation. 

\begin{table}[t]
\centering

\begin{tabular}{l c}
\toprule
\textbf{Metric} & \textbf{Cohen's $\kappa$} \\
\midrule
Semantic Entropy (ours) & \textbf{0.56} \\
Cosine similarity & 0.49 \\
Self-BLEU & 0.35 \\
Distinct-1 & 0.37 \\
Distinct-2 & 0.34 \\
\bottomrule
\end{tabular}
\vspace{-0.2cm}
\caption{Agreement between diversity metrics and human judgments of divergent creativity on MacGyver solutions.}
\label{tab:human_agreement}
\vspace{-0.6cm}
\end{table}

\paragraph{Semantic Entropy aligns with human judgments of divergent creativity.} To evaluate divergent creativity against human judgments, we conducted an annotation study on 50 MacGyver problems. Three LLMs (GPT-4o, Qwen-3-32B T, and Qwen-3-32B NT) were compared using pairwise human judgments, where three independent annotators selected the model exhibiting greater idea breadth for each comparison. Human preferences were aggregated by majority vote to produce a gold ranking per task. As shown in Table~\ref{tab:human_agreement}, SE achieves the highest agreement with human judgments ($\kappa$ = 0.56), outperforming cosine similarity, Self-BLEU, and Distinct-n. This shows that SE captures the semantic breadth humans associate with divergent creativity more faithfully than established surface-form diversity metrics. Full experimental details are provided in Appendix \ref{sec:se_human_validation}.

\begin{figure*}[t]
\vspace{-1.3cm}
\centering
\includegraphics[width=\textwidth]{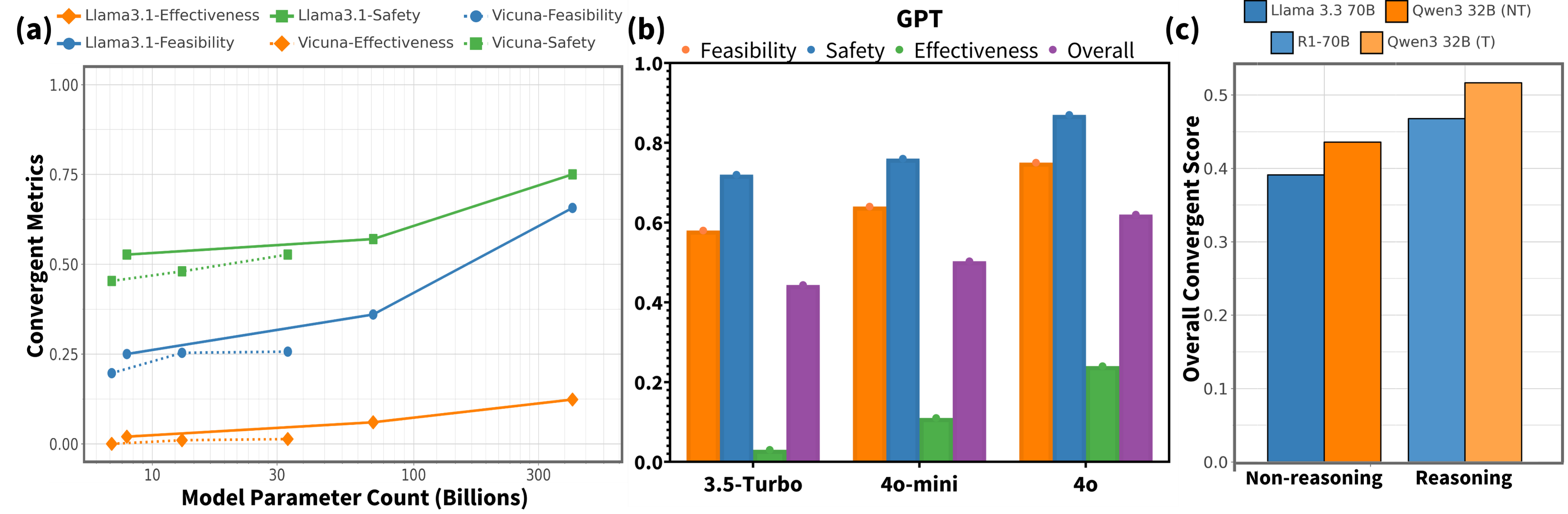}
\vspace{-0.8cm}
\caption{The impact of various parameters (left: model size, center: model recency, right: reasoning capabilities) on the convergent creativity of LLMs on the MacGyver dataset.}
\label{fig:convergent}
\vspace{-0.5cm}
\end{figure*}

\paragraph{Solution-level Semantic Entropy aligns with LLM-based novelty judgments.} We further examine whether solution-level SE aligns with novelty assessments of final solutions. We use an LLM-based pairwise novelty judge, first validated against human novelty rankings on 30 MacGyver solutions (Spearman's $\rho = 0.80$ with the majority-vote ground truth; see Appendix \ref{noveltyanalysis}). As shown in Fig. ~4d, the Spearman rank correlation between solution-level SE, computed as mean SE across all reasoning steps, and LLM-judged novelty rankings of final solutions is positive across models. This provides additional evidence that SE remains a meaningful indicator of creative breadth when aggregated at the solution level, and that greater local exploratory diversity during generation translates into more novel overall outputs.

\paragraph{Divergent creativity does not scale monotonically with model size or recency.} We analyze how divergent creativity varies with model scale and advancement using Semantic Entropy across multiple model families. As shown in Appendix~\ref{sec:addanalysis}, SE remains largely stable, and in some cases decreases, as models become larger or more recent, both across LLaMA generations (3, 3.1, 3.3; 8B to 405B) and Vicuna model sizes (7B to 33B). One plausible explanation is that contemporary training paradigms emphasize solution correctness, potentially constraining divergent output in larger models. Consistent with this observation, \citet{ruan2024EA} similarly report comparable levels of creative idea generation across less advanced and state-of-the-art models in scientific reasoning tasks.

\subsection{Convergent Creativity}

\paragraph{A retrieval-based multi-agent judge enables efficient and reliable evaluation of convergent creativity.} Our retrieval-based multi-agent framework achieves accuracy comparable to individual human annotators across both MacGyver and BookMIA (Table~\ref{tab:framework_performance}; annotation protocols in Table~\ref{sec:judge_eval}), and outperforms single-agent baselines and ChatEval on both domains. Human annotation for HypoGen was infeasible due to the specialized domain expertise required (Appendix \ref{hypogen}). By incorporating retrieval-based context selection and confidence-based stopping, the framework reduces computational cost by over 60\% relative to traditional multi-agent discussions (Appendix \ref{tab:compute}; \citealp{chan2023EA}), enabling large-scale evaluation. Overall, automated multi-agent judges can match human reliability across diverse creative domains while remaining practical for repeatable evaluation of open-ended, multi-criteria tasks.

\begin{table}[ht]
\centering
\setlength{\tabcolsep}{6pt}
\renewcommand{\arraystretch}{0.75}
\begin{tabular}{p{0.40\columnwidth} c c}
\toprule
Framework & \multicolumn{2}{c}{Accuracy} \\
\cmidrule(lr){2-3}
 & MacGyver & BookMIA \\
\midrule
\textbf{Baselines (GPT-4o)} \\
One-shot & 64.7\% & -- \\
CoT & 67.3\% & -- \\
Few-shot & 65.3\% & -- \\
Few-shot w/ CoT & 66.0\% & -- \\
ChatEval & 76.7\% & 73.3\% \\
\midrule
\textbf{Our framework} \\
GPT-4o-mini & 55.3\% & -- \\
GPT-4o & \textbf{84.7\%} & \textbf{83.0\%} \\
\midrule
\textbf{Human} \\
Annotator 1 & 82.7\% & 75.3\% \\
Annotator 2 & 84.7\% & 74.7\% \\
Annotator 3 & 81.3\% & 87.0\% \\
Annotator 4 & 80.0\% & -- \\
Annotator 5 & 81.3\% & -- \\
\bottomrule
\end{tabular}
\vspace{-0.3cm}
\caption{Performance of evaluation frameworks and human annotators on 50 MacGyver and 50 BookMIA solutions, against majority-vote gold standards (5 annotators for MacGyver, 3 for BookMIA).}
\label{tab:framework_performance}
\end{table}
\vspace{-0.2cm}

\begin{figure*}[t]
\vspace{-1.4cm}
\centering
\includegraphics[width=\textwidth]{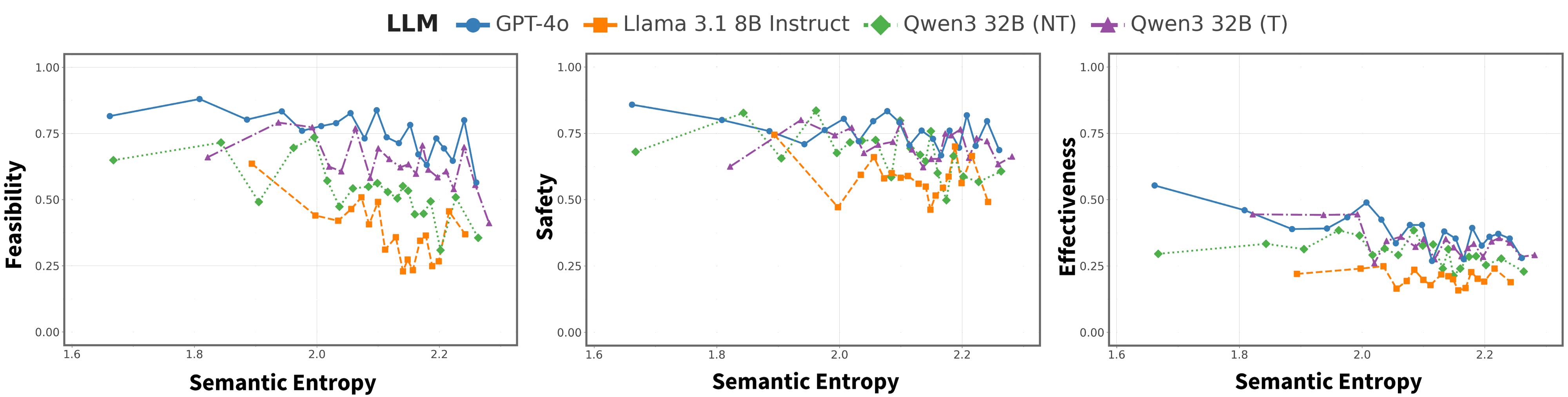}
\vspace{-0.8cm}
\caption{Semantic Entropy compared to different convergent creativity metrics (Y-axis) on the MacGyver dataset. Each point represents the mean Y value at the median X value of a unique set of 15 data points (fixed-interval binning). Similar trends were observed on the Hypogen and BookMIA datasets (see Appendix \ref{sec:dvc main}).}
\label{fig:sevsconvergent}
\vspace{-0.2cm}
\end{figure*}

\paragraph{Task fulfilment improves with model scale, recency, and reasoning.} We find that larger and more recent LLMs consistently achieve higher scores on convergent creativity metrics, reflecting task fulfilment. GPT-4o and LLaMA 3.1 70B outperform earlier or smaller counterparts (e.g., GPT-3.5, LLaMA 3.1 8B) across benchmarks (Fig.~5a,b). This trend aligns with findings that scaling and modern training strategies improve instruction following and structured reasoning. Additionally, reasoning models (e.g., R1-70B) outperform their non-reasoning base variants (e.g. Llama 3.3, Fig. 5c), highlighting the role of explicit reasoning mechanisms in satisfying complex, multi-criteria task requirements \citep{deepseekai2025deepseekr1incentivizingreasoningcapability}.

\subsection{Relationship Between Divergent and Convergent Creativity}
\label{results DvC}

\paragraph{Divergent and convergent creativity are empirically separable.} Spearman rank correlations between Semantic Entropy and convergent creativity metrics are consistently weak across all evaluated models and domains (Fig. \ref{fig:sevsconvergent}), indicating that higher SE does not systematically correspond to poorer task fulfilment, nor vice versa. Combined with our findings that divergent creativity does not scale with model size or recency while convergent creativity does, this suggests current training paradigms primarily improve structured reasoning without expanding creative exploration, and that the two dimensions may be independently optimizable. We note that convergent scores are produced by our LLM-based judge; while it aligns well with human annotations (Table \ref{tab:framework_performance}), these correlations should be interpreted accordingly.

\paragraph{Semantic Entropy captures divergent exploration rather than model error.} The weak correlation between SE and convergent metrics also serves as empirical validation of our core hypothesis. If SE in open-ended settings primarily reflected hallucination or model incorrectness, as in single-answer QA settings, one would expect a strong negative correlation with task fulfilment. The absence of such a relationship across all domains supports the interpretation that SE captures distributional breadth in generative reasoning rather than error.

\subsection{Additional Analyses}

Beyond the main findings above, we conducted further analyses on (1) the effect of temperature on convergent creativity, (2) the effect of sample size on Semantic Entropy, (3) the effect of step number on Semantic Entropy, and (4) the sensitivity of our framework's accuracy to varying confidence thresholds in appendix \ref{sec:addanalysis}. We also performed ablations on Semantic Entropy, including comparisons to naive entropy and alternative step-aggregation methods (min, max) in appendix \ref{sec:ablation}.

\begin{table*}[t]
\sisetup{round-mode=places, round-precision=2} 
    \vspace{-1cm}
    \centering
    \renewcommand{\arraystretch}{0.8} 

    \begin{adjustbox}{width=\linewidth}
        \begin{tabular}{
        l
        S[table-format=1.2]
        S[table-format=1.3]
        S[table-format=1.3]
        S[table-format=1.3]
        S[table-format=1.3]
        }
            \toprule
            {\textbf{Model}} & \multicolumn{1}{c}{\textbf{Divergent Creativity}} & \multicolumn{4}{c}{\textbf{Convergent Creativity}} \\
            \cmidrule(lr){2-2} \cmidrule(lr){3-6}
             & {\textbf{Semantic Entropy}} & {\textbf{Feasibility}} & {\textbf{Safety}} & {\textbf{Effectiveness}} & {\textbf{Overall}} \\
            \midrule
            Vicuna 7B & \textbf{2.19} & 0.20 & 0.45 & 0.00 & 0.217 \\
            Vicuna 13B & 1.96 & 0.25 & 0.48 & 0.01 & 0.248 \\
            Vicuna 33B & 2.17 & 0.26 & 0.53 & 0.01 & 0.257 \\
            \midrule
            Llama 3 70B Instruct & 2.10 & 0.39 & 0.65 & 0.02 & 0.356 \\
            \midrule
            Llama 3.1 8B Instruct & 2.13 & 0.25 & 0.53 & 0.02 & 0.266 \\
            Llama 3.1 70B Nemotron Instruct & \textbf{2.19} & 0.36 & 0.57 & 0.06 & 0.33 \\
            Llama 3.1 405B Instruct & 2.08 & 0.66 & 0.75 & 0.12 & 0.51 \\
            \midrule
            Llama 3.3 70B Instruct & 2.10 & 0.45 & 0.68 & 0.04 & 0.391 \\
            Deepseek R1 70B Distilled & 2.10 & 0.58 & 0.75 & 0.07 & 0.468 \\
            \midrule
            GPT 3.5 Turbo & 2.02 & 0.51 & 0.71 & 0.03 & 0.416 \\
            GPT 4o mini & 2.05 & 0.62 & 0.76 & 0.12 & 0.497 \\
            GPT 4o & 2.08 & \textbf{0.82} & \textbf{0.86} & \textbf{0.21} & \textbf{0.63} \\
            \midrule
            Qwen3 32B (Thinking) & 2.02 & 0.65 & 0.78 & 0.12 & 0.517 \\
            Qwen3 32B (Non-think) & 2.08 & 0.49 & 0.74 & 0.08 & 0.436 \\
            
            \bottomrule
        \end{tabular}
    \end{adjustbox}
    \vspace{-0.3cm}
    \caption{Performance of various LLMs on our benchmark using the MacGyver dataset.}
   \label{tab:macgyver performance}
   
\end{table*}
\begin{table*}[h!]
\sisetup{round-mode=places, round-precision=2} 

    \centering
    \vspace{-0.1cm}
    \renewcommand{\arraystretch}{0.8}
    \begin{adjustbox}{width=\linewidth}
        \begin{tabular}{
        l
        S[table-format=1.2]
        S[table-format=1.3]
        S[table-format=1.3]
        S[table-format=1.3]
        S[table-format=1.3]
        }
            \toprule
            {\textbf{Model}} & \multicolumn{1}{c}{\textbf{Divergent Creativity}} & \multicolumn{4}{c}{\textbf{Convergent Creativity}} \\
            \cmidrule(lr){2-2} \cmidrule(lr){3-6}
             & {\textbf{Semantic Entropy}} & {\textbf{Feasibility}} & {\textbf{Relevance}} & {\textbf{Scientific Accuracy}} & {\textbf{Overall}} \\
            \midrule
            GPT-4o & \textbf{2.07} & 0.28 & 0.61 & 0.17 & 0.353 \\
            Llama 3.1 8B Instruct & 2.04 & 0.21 & 0.56 & 0.12 & 0.293 \\
            Qwen3 32B (Thinking) & 1.72 & 0.41 & \textbf{0.78} & 0.21 & 0.467 \\
            Qwen3 32B (Non-think) & 1.66 & \textbf{0.50} & 0.76 & \textbf{0.26} & \textbf{0.51} \\
            \bottomrule
        \end{tabular}
    \end{adjustbox}
    \vspace{-0.3cm}
    \caption{Performance of various LLMs on our benchmark using the HypoGen dataset.}
    
    \label{tab:hypo_model_performance}

\end{table*}

\begin{table*}[h!]
\sisetup{round-mode=places, round-precision=2} 

    \centering
    \vspace{-0.1cm}
    \renewcommand{\arraystretch}{0.8}
    \begin{adjustbox}{width=\linewidth}
        \begin{tabular}{l S[table-format=1.2] *{4}{S[table-format=1.3]}}
            \toprule
            {\textbf{Model}} & \multicolumn{1}{c}{\textbf{Divergent Creativity}} & \multicolumn{4}{c}{\textbf{Convergent Creativity}} \\
            \cmidrule(lr){2-2} \cmidrule(lr){3-6}
             & {\textbf{Semantic Entropy}} & {\textbf{Coherence}} & {\textbf{Realism}} & {\textbf{Plot Completion}} & {\textbf{Overall}} \\
            \midrule
            GPT-4o & 2.17 & 0.36 & 0.40 & 0.23 & 0.33 \\
        Llama 3.1 8B Instruct & 1.89 & 0.03 & 0.04 & 0.03 & 0.03 \\
        Qwen3 32B (Thinking) & \textbf{2.19} & \textbf{0.50} & \textbf{0.41} & \textbf{0.52} & \textbf{0.48} \\
        Qwen3 32B (Non-think) & \textbf{2.19} & 0.36 & 0.44 & 0.35 & 0.38 \\
            \bottomrule
        \end{tabular}
    \end{adjustbox}
    \vspace{-0.3cm}
    \caption{Performance of various LLMs on our benchmark using the BookMIA dataset.}
    \label{tab:bookmia_model_performance}

\end{table*}

\section{Conclusion}

\paragraph{Key findings.} We introduce an automated, domain-general framework for evaluating LLM creativity, operationalizing divergent creativity through semantic entropy and convergent creativity through a retrieval-based multi-agent judge. Across three qualitatively distinct domains (MacGyver, HypoGen, BookMIA), semantic entropy reliably captures generative breadth, aligning with human judgments of novelty, the TTCT dimension of flexibility, and established surface-form diversity metrics, while remaining empirically separable from task fulfilment. Our retrieval-based judge matches individual human annotators on both domains where human validation was feasible, at over 60\% lower computational cost than prior multi-agent frameworks. Notably, divergent and convergent creativity scale differently with model size and recency: convergent performance improves with scale while divergent exploration does not, suggesting that current training paradigms primarily optimize correctness and structured reasoning rather than creative breadth.

\paragraph{Implications and future directions.} The empirical separability of divergent and convergent creativity has both interpretive and practical consequences. Interpretively, it supports treating creativity in LLMs as a multidimensional phenomenon rather than a single axis, and cautions against conflating task-fulfilment quality with creative capacity. Practically, it suggests the two dimensions may be independently optimizable: targeted interventions such as diversity-encouraging training objectives or sampling strategies could enhance creative breadth without sacrificing task fulfilment. More broadly, by decoupling measurement from the creative task itself, our framework provides a reproducible foundation for studying how architectural, training, and inference-time choices shape creative behavior in LLMs, supporting principled progress toward creative AI.

\section{Limitations}

Despite the demonstrated robustness and scalability of our proposed framework, several limitations merit consideration.

\paragraph{Computational Overhead.} The semantic entropy based evaluation, while effective in capturing novelty and diversity, remains computationally intensive. It requires the generation and clustering of multiple outputs per task, which may limit scalability when applied to large datasets or in environments with constrained computational resources.

\paragraph{Evaluation Limitations and Subjectivity.} Our current methodology assesses feasibility and task fulfilment using retrieval-based, context-sensitive LLM judgments. While this enables automated evaluation, it inherently relies on the subjective interpretation and prior knowledge encoded in the LLMs. As a result, unconventional or novel solutions that deviate from known patterns may be incorrectly deemed infeasible. Furthermore, the lack of real-world validation means that theoretically viable but unorthodox responses could be undervalued. Consequently, the framework may not fully capture the practical applicability or ingenuity of certain creative outputs.

\section*{Acknowledgements}

This research is supported by the Ministry of Education, Singapore, under its Academic Research Fund Tier 2 (MOE-T2EP20125-0002) and Tier 1 (RG22/24), National Research Foundation, Singapore under its National Large Language Models Funding Initiative (AISG Award No: AISG-NMLP-2024-001) and NTU Start Up Grant. Any opinions, findings and conclusions or recommendations expressed in this material are those of the author(s) and do not reflect the views of National Research Foundation, Singapore. The computational work for this article was partially performed on resources of the National Supercomputing Centre (NSCC), Singapore \url{https://www.nscc.sg}.

\bibliography{references}

\newpage
\appendix
\section*{Appendix}

\section{Model Selection} 
\label{appendixsth}
Our framework encompasses models of varying sizes, ages, and families.
The open-source models comprise 5 Llama models (\texttt{Llama-3.1-8B-Instruct,
Llama-3.1-Nemotron-70B-Instruct-HF, Llama-3.1-405B-Instruct, Llama-3-70B-Instruct, Llama-3.3-70B-Instruct}) \citep{grattafiori2024EA, wang2024EAhelpsteer} and 3 models from the Vicuna family (\texttt{vicuna-7b-v1.5, vicuna-13b-v1.5, vicuna-33b-v1.3}) \citep{chiang2023EA, zheng2023EA}. In addition, we also evaluate OpenAI\textquotesingle s \texttt{gpt-4o}, \texttt{gpt-3.5-turbo} and \texttt{gpt-4o-mini} closed-source models \citep{brown2020EA, openai2024EA}. Furthermore, we evaluate DeepSeek R1 70B Distilled \citep{deepseekai2025deepseekr1incentivizingreasoningcapability}, and Qwen3 32B \citep{qwen3} in both its thinking and non-thinking modes. Open-source models were obtained from Hugging Face.

\section{Code Availability}

Our code is available at the URL: \url{https://github.com/tanminsen/creativity-eval}. Our benchmark is intended exclusively for research purposes, and is not aimed for commercialisation, making it compatible with original access conditions. 

\section{Semantic Entropy}

\subsection{Estimation and Length Normalization}

In practice, not all possible responses from all possible semantic
classes can be sampled from the LLM to compute semantic entropy.
Therefore, we follow Farquhar et al. (2024) and estimate the semantic
entropy using a Rao-Blackwellized Monte Carlo integration over the
semantic classes \(C\):

\vspace{-0.3cm}
\begin{equation}
H(x) \approx - \sum_{i = 1}^{|C|}P(C_{i}|x)\log P(C_{i}|x)\
\end{equation}

Where \(P(C_{i}|x) = \frac{P(c_{i}|x)}{{\sum_{}^{}}_{c}P(c|x)}\). This
normalises the semantic class probabilities by taking the semantic
classes as a categorical distribution.

\noindent\paragraph{} To account for disparities in output sequence length, which inherently
affect the combined likelihood, we employ length normalization during
the computation of log-probabilities for generated sequences. This
procedure addresses the principle of conditional independence in token
probability distributions \citep{malininandgales2021}, wherein the
probability of a sequence diminishes exponentially with its length.
Consequently, without normalization, the negative log-probability
increases linearly with sequence length, leading to a bias where longer
sequences disproportionately contribute to the measured entropy.
Therefore, we calculate the joint log-probability of a sequence as the mean of the sequence instead of the sum:

\vspace{-0.4cm}
\begin{equation}
\log P(s|x) = \frac{1}{N}\sum_{i = 1}^{N}\log P(t_{i}|t_{< i},x)\ \
\end{equation}

\subsection{Sampling Solutions from LLMs}

When sampling generations, we set a default temperature of 1.0 (unless
stated otherwise), with nucleus sampling (top\_p = 0.9).

\subsection{Semantic Clustering Entailment}\label{entailmentdetails}

\subsubsection{Entailment Models}

To guide our implementation, we conducted an empirical comparison of several popular entailment models, including GPT-4o (zero-shot) and multiple NLI models. Performance was validated on 55 manually annotated sentence pairs from the Macgyver Dataset, with three independent human annotators providing the ground truth (inter-annotator agreement: Cohen’s Kappa = 0.61). Table \ref{tab:entailmentperfs} reports accuracy relative to this ground truth. 

Detailed NLI model URLs from Hugging Face: 

\begin{itemize}
    \setlength{\itemsep}{0pt}
    \setlength{\parskip}{0pt}
    \setlength{\parsep}{0pt}
    \setlength{\topsep}{2pt}
    \item tasksource/deberta-base-long-nli
    \item facebook/bart-large-mnli
    \item MoritzLaurer/DeBERTa-v3-large-mnli-fever-anli-ling-wanli
    \item FacebookAI/roberta-large-mnli
    \item NDugar/v3-Large-mnli
\end{itemize}

\begin{table}[!htbp]
\centering
\setlength{\tabcolsep}{6pt}
\begin{tabular}{@{}l r@{}}
\toprule
Model & Accuracy \\
\midrule
DeBERTa-Base-Long-NLI & \textbf{78.1\%} \\
GPT-4o & 72.7\% \\
DeBERTa-v3-Large (GLUE MNLI) & 67.3\% \\
Bart-Large-MNLI & 52.7\% \\
DeBERTa-v3-Large & 52.7\% \\
RoBERTa-Large-MNLI & 47.2\% \\
\bottomrule
\end{tabular}
\vspace{-0.2cm}
\caption{Entailment models.}
\label{tab:entailmentperfs}
\end{table}

\subsubsection{Creation of Entailment Ground Truth Dataset}
To evaluate the entailment performance of various LLMs, 3 human annotators were approached. Each was given 55 randomly sampled pairs of steps generated by either Llama 3.1 8B Instruct of GPT-4o, as part of solutions to problems from the MacGyver dataset, and were asked to provide binary verdicts on whether entailment was present in each pair. The final ground truth was produced via majority voting. Kappa coefficients are presented in table \ref{tab:kappa_matrix_entailment}. The proportion of positive and negative entialments in the ground truth dataset are provided in table \ref{tab:entailment_ground_truth_labels}. 
\begin{table}[ht]
\centering

\begin{tabular}{l S[table-format=1.6] S[table-format=1.6] S[table-format=1.6]}
\toprule
Annotator       & {\textbf{1}} & {\textbf{2}} & {\textbf{3}} \\
\midrule
\textbf{1} & NA            & 0.60     & 0.74      \\
\textbf{2} & 0.60     & NA             & 0.49     \\
\textbf{3} & 0.74      & 0.49     & NA             \\
\bottomrule
\end{tabular}
\caption{Inter-Annotator Agreement Matrix (Cohen's Kappa) for Entailment Ground Truth}
\label{tab:kappa_matrix_entailment}
\end{table}

\begin{table}[h!]
  \centering
  
  \begin{tabular}{
    l
    S[table-format=2.0] 
  }
    \toprule
    \textbf{Verdict} & {\textbf{Count}} \\
    \midrule
    Positive & 26 \\
    Negative & 29 \\
    \bottomrule
  \end{tabular}
  \caption{Distribution of labels in the Entailment Ground Truth dataset.}
  \label{tab:entailment_ground_truth_labels}
\end{table}

\subsubsection{Clustering Algorithm}

\paragraph{}We use \texttt{tasksource/deberta-base-long-nli} as our DeBERTa model to cluster
samples into semantic classes. We selected it for our framework because it achieved the highest accuracy on the MacGyver dataset, which we considered essential to ensure the reliability and consistency of downstream analyses for our specific domain case. At the same time, the framework remains agnostic to the choice of entailment model: researchers are free to substitute stronger or more efficient encoders as they become available. 

\paragraph{}The details for the greedy entailment
algorithm, retrieved from \citet{SE}, are as follows:

\paragraph{}For each sample \(s_{a}\), we obtain the bidirectional entailment
between it and a sample from an existing semantic class \(C_{k}\) ; if
entailment is found, \(s_{a}\) is appended to the class; if its semantic
meaning differs from those of all existing classes, it forms its own
class. Iterating through all samples \(s_{1}...s_{n}\), we obtain the
set of semantic classes wherein the samples are fully clustered.

\paragraph{}In other words, if two outputs \(s_{a}\) and \(s_{b}\) mutually entail one
another, they are considered part of the same semantic class. 
For each sample \(s_{a}\), we obtain the bidirectional entailment
between it and a sample from an existing semantic class \(C_{k}\) ; if
entailment is found, \(s_{a}\) is appended to the class; if its semantic
meaning differs from those of all existing classes, it forms its own
class.

\subsection{Semantic Entropy Captures Diversity}
\label{seanalysis}

\paragraph{}Existing literature has proposed various means of quantifying the diversity or semantic consistency of a set of LLM generations in order to probe its creativity. This includes cosine similarity \citep{cossim1, yang2025enhancingsemanticconsistencylarge}, the Self-BLEU metric \citep{selfbleu}, and distinct-n scores \citep{distinctn}. By computing the aforementioned metrics for samples generated from our benchmark, we explore the relationship between them and semantic entropy, as shown in tables \ref{tab:se_diversity_vertical},  \ref{tab:se_diversity_vertical2} and \ref{tab:se_diversity_vertical3}. 

\paragraph{}The variability in correlations between Semantic Entropy (SE) and other diversity metrics across Tables 9–11 is expected and does not indicate instability of the measure. The compared metrics primarily capture surface-form or lexical diversity, whereas SE measures semantic dispersion across distinct idea classes. These dimensions of diversity are inherently different: classical creativity theory similarly distinguishes semantic flexibility from fluency or surface-level variation.

\paragraph{}Because each dataset emphasizes different forms of creative variation, surface-based metrics respond inconsistently across domains. In contrast, SE consistently captures semantic breadth by explicitly grouping generations according to underlying meaning rather than token overlap or embedding proximity.

\paragraph{}In this sense, SE provides an orthogonal measure of divergent creativity rather than replicating existing diversity metrics. If SE were highly correlated with n-gram or token-level measures across all domains, it would suggest redundancy rather than a meaningful semantic signal. The observed pattern instead supports SE as a distinct and principled indicator of semantic exploration in generative reasoning.

\begin{table*}[ht!]
  \centering

  \begin{tabular}{llrr}
    \toprule
    Metric    & Model            & $\rho$       & p-value                  \\
    \midrule
    \textbf{Average Cosine Similarity} \\
      & GPT-4o                  & $-0.436$     & $6.72\times10^{-58}$     \\
      & Llama 3.1 8B Instruct       & $-0.048$     & $3.95\times10^{-2}$      \\
      & Qwen3 32B Non-thinking           & $-0.456$     & $3.18\times10^{-125}$     \\
      & Qwen3 32B Thinking       & $-0.246$     & $6.09\times10^{-22}$     \\
    \textbf{Self-BLEU} \\
      & GPT-4o                  & $-0.374$     & $3.25\times10^{-43}$     \\
      & Llama 3.1 8B Instruct       & $0.119$      & $3.58\times10^{-7}$      \\
      & Qwen3 32B Non-thinking           & $-0.475$     & $9.81\times10^{-137}$     \\
      & Qwen3 32B Thinking       & $-0.149$     & $7.44\times10^{-9}$      \\
    \textbf{Distinct-1} \\
      & GPT-4o                  & $0.285$      & $3.23\times10^{-21}$     \\
      & Llama 3.1 8B Instruct       & $-0.238$     & $5.19\times10^{-25}$     \\
      & Qwen3 32B Non-thinking           & $0.382$      & $4.40\times10^{-85}$     \\
      & Qwen3 32B Thinking       & $0.048$      & $6.49\times10^{-2}$      \\
    \textbf{Distinct-2} \\
      & GPT-4o                  & $0.349$      & $5.35\times10^{-36}$     \\
      & Llama 3.1 8B Instruct       & $-0.140$     & $1.92\times10^{-9}$      \\
      & Qwen3 32B Non-thinking           & $0.452$      & $1.56\times10^{-122}$     \\
      & Qwen3 32B Thinking       & $0.118$      & $5.39\times10^{-6}$      \\
    \bottomrule
  \end{tabular}
  \caption{Spearman correlation ($\rho$) between semantic entropy and diversity metrics across models for the MacGyver dataset.}\label{tab:se_diversity_vertical}
\end{table*}
\begin{table*}[h!]
  \centering

  \begin{tabular}{llrr}
    \toprule
    Metric       & Model                         & $\rho$     & p-value               \\
    \midrule
    \textbf{Average Cosine Similarity} \\
      & GPT-4o                        & $-0.396$   & $1.34\times10^{-108}$ \\
      & Llama 3.1 8B Instruct         & $-0.343$   & $9.41\times10^{-92}$  \\
      & Qwen3 32B Non-thinking        & $-0.463$   & $2.40\times10^{-158}$ \\
      & Qwen3 32B Thinking            & $-0.564$   & $4.11\times10^{-181}$ \\
    \textbf{Self-BLEU} \\
      & GPT-4o                        & $-0.346$   & $2.01\times10^{-81}$  \\
      & Llama 3.1 8B Instruct         & $-0.174$   & $2.02\times10^{-21}$  \\
      & Qwen3 32B Non-thinking        & $-0.440$   & $1.89\times10^{-141}$ \\
      & Qwen3 32B Thinking            & $-0.438$   & $7.30\times10^{-102}$ \\
    \textbf{Distinct-1} \\
      & GPT-4o                        & $0.415$    & $9.87\times10^{-120}$ \\
      & Llama 3.1 8B Instruct         & $0.180$    & $9.27\times10^{-23}$  \\
      & Qwen3 32B Non-thinking        & $0.516$    & $2.03\times10^{-202}$ \\
      & Qwen3 32B Thinking            & $0.570$    & $2.28\times10^{-186}$ \\
    \textbf{Distinct-2} \\
      & GPT-4o                        & $0.405$    & $7.93\times10^{-114}$ \\
      & Llama 3.1 8B Instruct         & $0.206$    & $1.95\times10^{-29}$  \\
      & Qwen3 32B Non-thinking        & $0.496$    & $3.38\times10^{-185}$ \\
      & Qwen3 32B Thinking            & $0.529$    & $5.49\times10^{-156}$ \\
    \bottomrule
  \end{tabular}
  \caption{Spearman correlation ($\rho$) between semantic entropy and diversity metrics for the HypoGen dataset.}\label{tab:se_diversity_vertical2}
\end{table*}

\begin{table*}[h!]
  \centering

  \begin{tabular}{llrr}
    \toprule
    Metric       & Model                         & $\rho$     & p-value               \\
    \midrule
    \textbf{Average Cosine Similarity} \\
      & GPT-4o                        & $-0.261$ & $6.52\times10^{-42}$  \\
      & Llama 3.1 8B Instruct         & $-0.112$ & $0.00714$             \\
      & Qwen3 32B Non-thinking        & $-0.119$ & $7.84\times10^{-11}$  \\
      & Qwen3 32B Thinking            & $-0.022$& $0.22966$             \\
    \addlinespace
    \textbf{Self-BLEU} \\
      & GPT-4o                        & $-0.463$   & $< 1 \times 10^{-200}$ \\
      & Llama 3.1 8B Instruct         & $-0.448$   & $5.83\times10^{-30}$  \\
      & Qwen3 32B Non-thinking        & $-0.372$   & $6.65\times10^{-98}$  \\
      & Qwen3 32B Thinking            & $-0.058$   & $0.00197$             \\
    \addlinespace
    \textbf{Distinct-1} \\
      & GPT-4o                        & $0.392$   & $8.02\times10^{-97}$  \\
      & Llama 3.1 8B Instruct         & $-0.006$    & $0.885$             \\
      & Qwen3 32B Non-thinking        & $0.232$   & $1.48\times10^{-37}$  \\
      & Qwen3 32B Thinking            & $0.004$    & $0.82144$             \\
    \addlinespace
    \textbf{Distinct-2} \\
      & GPT-4o                        & $0.459$   & $< 1 \times 10^{-200}$ \\
      & Llama 3.1 8B Instruct         & $0.173$   & $< 1 \times 10^{-200}$ \\
      & Qwen3 32B Non-thinking        & $0.321$   & $< 1 \times 10^{-200}$ \\
      & Qwen3 32B Thinking            & $-0.0208$   & $0.264$             \\
    \bottomrule
  \end{tabular}
  \caption{Spearman correlation ($\rho$) between semantic entropy and diversity metrics for the BookMIA dataset.}\label{tab:se_diversity_vertical3}
\end{table*}

\newpage

\subsection{Semantic Entropy Captures Novelty}

\begin{table}[h!]
  \centering

  \begin{tabular}{
    l
    S[table-format=-1.5]
    S[table-format=-1.5]
  }
    \toprule
    \textbf{Model} & {\textbf{MacGyver}} & {\textbf{BookMIA}} \\
    \midrule
    Llama 3.1 8B   & 0.310 & 0.307 \\
    Qwen3 32B NT & 0.456  & 0.478 \\
    Qwen3 32B T  & 0.469 & 0.229 \\
    GPT-4o    & 0.421 & 0.432 \\
    \bottomrule
  \end{tabular}
  \vspace{-0.1cm}
  \caption{Spearman rank correlation between Novelty and Semantic Entropy on the MacGyver and BookMIA benchmarks, for different LLMs.}
  \label{tab:novelty_corr}
\end{table}

Table \ref{tab:novelty_corr} reveals a consistent and significant positive correlation between semantic entropy and novelty across all tested LLMs and datasets. It provides robust quantitative evidence that an increase in semantic entropy corresponds directly to an increase in the novelty of the generated solution. This finding further validates the use of semantic entropy as an effective and comprehensive proxy for novelty, a critical dimension of creative output that is otherwise difficult to measure at scale. The strength and consistency of this relationship across different model architectures and datasets firmly establish its generalizability. 

\subsection{Comparison to Existing Creativity Frameworks}

A popular and established framework to evaluate human creativity is the Torrance Tests of Creative Thinking (TTCT) \cite{torranceND}. In this section, we compare it against our benchmark, and highlight why our benchmark is more applicable and suited for evaluating LLM creativity. 

The TTCT consists of 4 metrics: \textbf{originality, flexibility, fluency and elaboration}. 

Firstly, we specifically address \textbf{originality} by adding an LLM novelty judge to our evaluation, in addition to our semantic entropy (SE) metric. We first validated this judge by comparing its novelty ratings to those from human annotators and found high agreement. Using this setup, we showed that SE is strongly correlated with LLM-assessed novelty scores on our datasets. This directly demonstrates that SE robustly captures the originality aspect of creativity, as intended by TTCT.
    
Next, \textbf{flexibility} is measured through the diversity of semantic classes produced by the model for each problem. As already shown in our original paper, we included a graph illustrating a strong positive correlation between SE and the number of unique semantic classes generated. This provides quantitative evidence that our framework reflects flexibility in the TTCT sense, by capturing the range of different categories of solutions produced by the model.
    
In addition, we recognize that \textbf{fluency} — the sheer number of ideas produced — is a key component of TTCT’s evaluation of human creativity, particularly because, in human-administered tests, generation protocols are tightly standardized and ideation is effortful. For LLMs, however, fluency is governed by sampling parameters and can be trivially increased or decreased, making it less indicative of genuine creative ability in automated settings. Thus, we do not foreground fluency as a core metric in our framework, but acknowledge its value in structured human creativity tasks.*

Finally, while \textbf{elaboration} — the detail and development of ideas — is valuable in human TTCT tasks (where added depth reflects genuine effort and cognitive engagement), we do not account for elaboration as a core metric in our framework. In LLMs, elaboration can be easily manipulated by prompting for longer or more detailed responses, meaning that output length is decoupled from substantive creativity. Instead, we focus on task fulfilment through our convergent creativity evaluation, which provides a more relevant and robust assessment of whether a model’s response meets the requirements and constraints of the task.

\subsection{Evaluation of Solution Novelty}
\label{noveltyanalysis}
\subsubsection{Creation of Ground Truth Dataset}
We had 5 human annotators rank a set of 30 problem-solution pairs from the MacGyver dataset based on their novelty, and compared their rankings, finding moderate agreement between them. The golden ground truth was obtained by taking the average ranking of problem-solution pair by the 5 annotations. The inter-annotator spearman rank correlation is shown below in table \ref{tab:spearmannovelty}. Owing to general agreement between annotators, the ground truth for novelty is sufficiently robust.  

\begin{table}[h!]
\centering

\begin{tabular}{cccccc}
\toprule
Annotator & 1 & 2 & 3 & 4 & 5 \\
\midrule
1 & NA & 0.34 & 0.50 & 0.55 & 0.57 \\
2 & 0.34 & NA & 0.33 & 0.57 & 0.50 \\
3 & 0.50 & 0.33 & NA & 0.49 & 0.53 \\
4 & 0.55 & 0.57 & 0.49 & NA & 0.55 \\
5 & 0.57 & 0.50 & 0.53 & 0.55 & NA \\
\bottomrule
\end{tabular}
\caption{Average Pairwise Spearman Rank Correlation for Annotator Agreement}
\label{tab:spearmannovelty}
\end{table} 

The 30 problem-solution pairs for the ground truth are sampled from GPT-4o, Llama 3.1 8B Instruct, and Vicuna 7B. 

\subsubsection{Automated Novelty Evaluation}

To automate novelty evaluation, we use an LLM Judge to determine novelty using pairwise comparisons between problem-solution pairs, integrated into a bubble sort algorithm. To compare its performance to human annotators, it evaluated the 30 problem-solution pairs in the ground truth dataset. 

\begin{figure}
    \centering
    \includegraphics[width=0.45\textwidth]{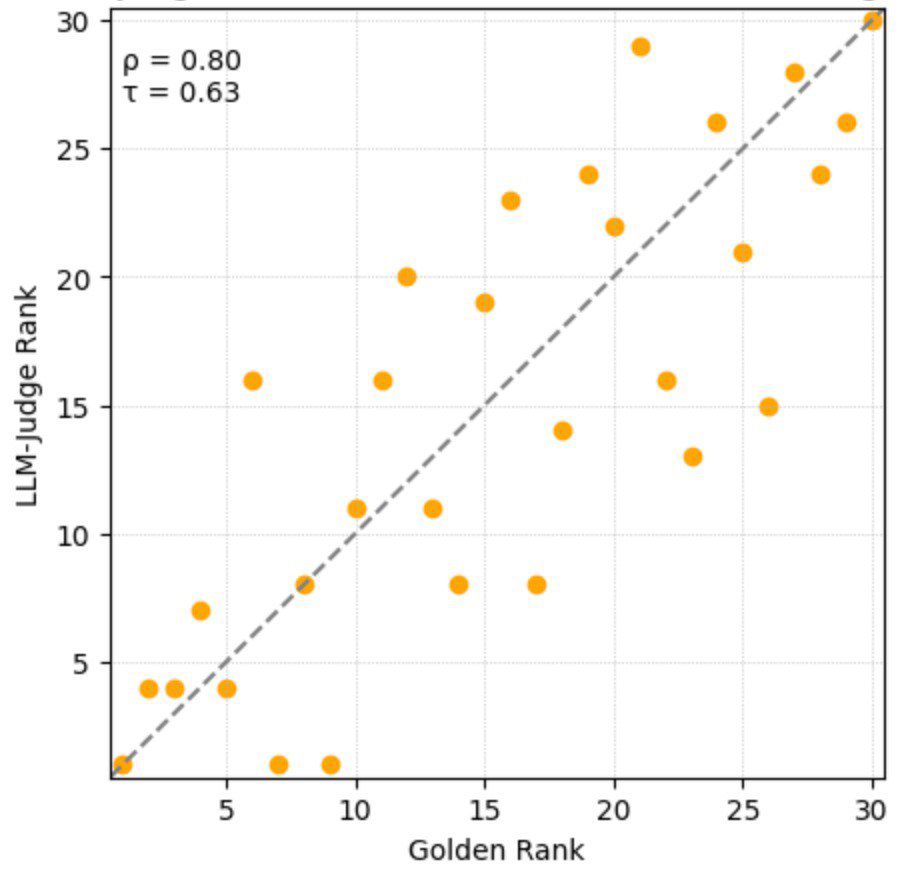}
    \vspace{-0.2cm}
    \caption{The correlation between LLMJudge novelty rankings and the ground truth ranks. }
    \label{fig:llmjnovelty}
    \vspace{-0.2cm}
\end{figure}

As shown in Fig. \ref{fig:llmjnovelty}, the LLM Judge used for pairwise novelty has strong agreement (Spearman $\rho$ of 0.80 and Kendall's $\tau$ of 0.63) with human annotation, and thus can reliably serve as an automated method for gauging the novelty of LLM responses. 
\subsection{Human Validation and Sampling Design Analysis for Semantic Entropy}
\label{sec:se_human_validation}

\paragraph{Experimental Setup.}
To examine the effect of decoding strategy on SE, we conducted a human annotation study on 50 MacGyver problems. Three LLMs were evaluated: GPT-4o, Qwen-3-32B (Thinking), and Qwen-3-32B (Non-Thinking). 

We compare two decoding regimes under matched sampling conditions:

\noindent\textbf{Condition A: Greedy decoding (default framework).}
At each reasoning step, we sample $k = 10$ candidate continuations to compute step-level SE. The next step is selected via greedy decoding, yielding a single solution trajectory and one SE score per model per question.

\noindent\textbf{Condition B: Multi-path decoding (5-path variant).}
At the first reasoning step, we sample $k = 10$ continuations and expand five independent solution trajectories. SE is computed along each trajectory and averaged to produce a single SE score per model per question. Only the first branching step is expanded, avoiding exponential growth.

This produces two SE estimates per model per question, enabling direct comparison between greedy and multi-path decoding.

\paragraph{Human Annotation Protocol.}
To obtain a human-grounded reference for divergent creativity, we recruited three independent annotators. For each MacGyver problem, each model produced five full solution paths (from Condition B). Annotators performed pairwise comparisons between models (three comparisons per question), selecting the model exhibiting greater divergent creativity, defined as broader and more varied ideation. Pairwise preferences were aggregated by majority vote to produce a single gold ranking per question. Agreement between automated metrics and human rankings was evaluated using Cohen’s $\kappa$.

\begin{table}[t]
\centering
\small
\setlength{\tabcolsep}{6pt}
\renewcommand{\arraystretch}{1.05}
\begin{tabular}{l c}
\toprule
\textbf{Metric} & \textbf{Cohen’s $\kappa$} \\
\midrule
Semantic Entropy (Greedy decoding) & \textbf{0.56} \\
Semantic Entropy (Multi-path decoding) & 0.48 \\
Cosine similarity & 0.49 \\
Self-BLEU & 0.35 \\
Distinct-1 & 0.37 \\
Distinct-2 & 0.34 \\
\bottomrule
\end{tabular}
\vspace{-0.1cm}
\caption{Agreement between human pairwise judgments of divergent creativity and automated diversity metrics on 50 MacGyver problems. Semantic Entropy computed under greedy decoding achieves the highest agreement.}
\label{tab:human_agreement_full}
\end{table}

\paragraph{Results.}
Table~\ref{tab:human_agreement_full} reports agreement between human rankings and various diversity metrics. Semantic Entropy under greedy decoding achieves the highest agreement with human judgments ($\kappa = 0.56$), outperforming both the multi-path SE variant and all surface-form diversity metrics, including cosine similarity, Self-BLEU, and Distinct-$n$.

\paragraph{Implications for Sampling Design.}
Despite concerns that greedy decoding may suppress divergent creativity, SE computed under greedy decoding aligns more strongly with human judgments than SE computed from multiple expanded trajectories. This suggests that greedy decoding does not obscure the divergent signal extracted from step-wise sampling. Instead, constraining the forward path appears to preserve discriminability between models by reflecting their intrinsic distribution over semantically distinct ideas.

A plausible explanation is that expanding multiple full trajectories may cause different models to converge toward similar diversity levels, reducing contrast in aggregate diversity measures.

\paragraph{Implications for Semantic Entropy.}
Across all evaluated metrics, Semantic Entropy under greedy decoding shows the strongest alignment with human judgments of divergent creativity. This provides additional empirical validation that SE captures semantic breadth rather than surface variation or model error, supporting its use as a robust, human-aligned measure of divergent creativity in open-ended generative tasks.

\subsection{Ablation Studies}\label{sec:ablation}
\subsubsection{Naive Entropy}

To further validate the added value of semantic entropy in quantifying divergent creativity, we conducted an ablation study comparing our semantic entropy (SE) with naïve sequence-level entropy. We computed Spearman correlations against human novelty rankings of the solutions (reported in table \ref{tab:entropy_comparison}) on the MacGyver and BookMIA datasets across four models (Llama 3.1 8B, Qwen 3 32B Thinking, Qwen3 32B Non-thinking, GPT-4o). These results indicate that semantic entropy achieves substantially stronger alignment with human judgments compared to naïve entropy, supporting its effectiveness for capturing meaningful diversity rather than surface-level token variation. 

\subsubsection{Semantic Entropy Aggregation Method}
\label{SEaggregation}

To validate that our approach reliably reflects overall creativity, we tested several aggregation strategies—arithmetic mean, minimum, and maximum—and computed Spearman correlations with human novelty rankings of the solutions. Results are presented in table \ref{tab:novelty_correlation_combined}. These results show that the arithmetic mean consistently outperformed the min and max aggregations, which were less stable and more sensitive to individual steps. We selected the mean because it accounts for the novelty present across all steps, providing a balanced and interpretable summary of diversity. Based on this evidence, we retained it as a robust default. 

\sisetup{round-mode=places, round-precision=2}

\begin{table}[t]
\sisetup{round-mode=places, round-precision=2} 
\centering
\small
\begin{tabular}{lcc}
\toprule

\multicolumn{3}{c}{\textbf{MacGyver}} \\
\cmidrule(lr){1-3}
Model & Na\"ive Entropy & Semantic Entropy \\
\midrule
LLaMA 3.1 8B    & -0.312 & -0.310 \\
Qwen3 32B NT & -0.234 & -0.456 \\
Qwen3 32B T    & -0.039 & -0.469 \\
GPT-4o                &  0.123 & -0.421 \\
\midrule

\multicolumn{3}{c}{\textbf{BookMIA}} \\
\cmidrule(lr){1-3}
Model & Na\"ive Entropy & Semantic Entropy \\
\midrule
LLaMA 3.1 8B   &  0.181 & -0.307 \\
Qwen3 32B NT & -0.297 & -0.478 \\
Qwen3 32B T    &  0.519 & -0.229 \\
GPT-4o                & -0.400 & -0.432 \\

\bottomrule
\end{tabular}
\caption{Spearman correlations between generation novelty and Na\"ive vs.\ Semantic Entropy across datasets.}
\label{tab:entropy_comparison}
\end{table}

\begin{table}[t]
\centering
\small
\sisetup{round-mode=places, round-precision=2}

\begin{tabular}{ll
  S[table-format=-1.2]
  S[table-format=-1.2]
  S[table-format=-1.2]
}
\toprule
Dataset & Model & {Min} & {Max} & {Mean} \\
\midrule

\multirow{5}{*}{\textbf{MacGyver}}
& LLaMA 3.1 8B   & -0.03 & -0.49 & -0.31 \\
& Qwen3 32B NT & -0.39 & -0.28 & -0.47 \\
& Qwen3 32B TT     & -0.32 & -0.12 & -0.46 \\
& GPT-4o                & -0.34 & -0.03 & -0.42 \\
& \textbf{Average}      & -0.27 & -0.23 & -0.41 \\
\midrule

\multirow{5}{*}{\textbf{BookMIA}}
& LLaMA 3.1 8B   & -0.26 & -0.31 & -0.31 \\
& Qwen3 32B NT & -0.41 & -0.39 & -0.48 \\
& Qwen3 32B T     & -0.16 & -0.38 & -0.23 \\
& GPT-4o                & -0.36 & -0.07 & -0.43 \\
& \textbf{Average}      & -0.30 & -0.29 & -0.36 \\

\bottomrule
\end{tabular}

\caption{Spearman correlation between novelty judgments and different semantic entropy aggregation strategies (minimum, maximum, arithmetic mean) across datasets. Higher values indicate stronger alignment with novelty.}
\label{tab:novelty_correlation_combined}
\end{table}

\section{Retrieval-based LLM Discussion Framework}

We use \texttt{dunzhang/stella\textunderscore en\textunderscore 1.5B\textunderscore v5} as our embedding model for the retrieval-based evaluation framework, and use a ChromaDB database to store the fragment embeddings. 
We set \(j = 4, k = 5, l = 8\) with confidence threshold \(T=0.5\). The agents were prompted to limit their responses to a maximum of 150 words. 

\subsection{Metrics}
\label{llmjmetrics}

\paragraph{MacGyver.} \textbf{Feasibility}: whether a solution is practical and can be realistically implemented. \textbf{Safety}: the potential for harm or risks associated with the solution, ensuring that it adheres to ethical and practical guidelines. \textbf{Effectiveness}: how well the solution achieves the desired outcome, focusing on efficiency and accuracy.

\paragraph{HypoGen.} Inspired by the dataset itself and Google's AI Co-scientist \citep{oneill2025sparkssciencehypothesisgeneration, gottweis2025aicoscientist}. \textbf{Feasibility}: the solution and reasoning chain is practical and likely to succeed. \textbf{Relevance}: the generated solution must precisely align with the research goals, preferences and constraints defined by the problem (bit and flip). \textbf{Scientific Accuracy}: the approaches, concepts, measurements, and models mentioned in the solution correctly represent the true nature or behavior of the phenomenon under investigation.

\paragraph{BookMIA.} Previously shown to be important to narrative quality \citep{yi2025scorestorycoherenceretrieval}. \textbf{Narrative Coherence}: the logic of the story is maintained - character behavior and plot development is consistent. It should avoid internal contradictions, and flow logically from its beginning to its conclusion. \textbf{Emotional and Psychological Realism}: the story can maintain consistent and believable emotional states and character behaviors, ensuring character consistency throughout the narrative. \textbf{Plot Completion}: the story successfully and logically progresses to the ending sentences provided, effectively linking the story's events to its intended conclusion

\subsection{Implementation Details}\label{sec:implementdetails}
\normalsize The framework organises structured discussions among three LLM agents, each with distinct roles: 
\begin{itemize}
    \item The \textbf{Problem Analyst} (PA) explores problem properties.
    \item The \textbf{Solution Analyst} (SA) assesses solutions.
    \item The \textbf{Criterion Analyst} (CA) refines criteria definitions.
\end{itemize}

\noindent We denote the set of analysts as $a \in \{\text{PA}, \text{SA}, \text{CA}\}$.

The problem $P$, solution $S$ and criterion $C_i$, where the criterion is within the set of all criteria $C$ to be evaluated, are jointly represented by $Info$. 

The process involves three phases: Initialization, Discussion, and Verdict.

\subsubsection{Fragments}
Each agent generates insights as structured information pieces called fragments, $F_i$. Fragments are stored in a database $D$ with their embeddings $E(F_i)$. Agents retrieve the $n$ most relevant fragments using a query $Q$, based on cosine similarity between $E(Q)$ and $E(F_i)$:

\vspace{-0.3cm}
\begin{equation}
    \begin{split}
        & \text{GET}(Q, n) = \text{Top-}n(\text{Sim}(\mathcal{E}(Q), \mathcal{E}(F_i))) \\
        & \text{where } F_i \in D
    \end{split}
\end{equation}

\subsubsection{Initialization}
Analysts ($J_a$ for $a \in \{\text{PA}, \text{SA}, \text{CA}\}$) generate initial insights about problem, solution, and criteria $C = (C_1, C_2, C_3)$ with definitions. The background information $(P, S, C_i)$ is denoted as $Info$. Parameters $k$, $j$, and $l$ define the number of fragments retrieved for discussion, scoring, and verdict phases, respectively.

\subsubsection{Discussion}
The core of our framework is structured discussion between analysts after all 3 analysts ($J_a$ for $a \in \{\text{PA}, \text{SA}, \text{CA}\}$) generate initial insights given $Info$, and is depicted in the equation below. 
Specifically, analysts provide:
\begin{enumerate}
    \item Answers to questions from other analysts $R^\text{response}_a$.
    \item General opinions $R^\text{opinion}_a$.
    \item Clarifying questions to other analysts $q^\text{new}_a$.
\end{enumerate}

\vspace{-0.4cm}

\begin{equation}
    \vspace{0.2cm}
    \begin{split}
        &(R^\text{questions}_a, R^\text{opinion}_a, q^\text{new}_a)  \\ 
        &= J_a(q^\text{others}_a, \text{GET}(Q_a \oplus q^\text{others}_a, k), Info).
    \end{split}
    \vspace{0.3cm}
\end{equation}

\noindent The analysts extract relevant fragments using predefined role-specific queries
Qa, and questions from other analysts. Their generated insights are stored in
the database.

\paragraph{Confidence Scoring.}
At the end of each round $r$, analysts assign confidence scores $C_a^{(r)}$ reflecting judgment reliability. If the mean score $C^{(r)}$ exceeds a threshold $T$, the discussion stops early as analysts are confident in their judgement. Else, the discussion continues (up to two rounds):

\begin{equation}
    C_a^{(r)} = J_a(\text{GET}(Q_a, j), Info).
\end{equation}

\noindent The threshold $T$ is a hyperparameter which has been investigated; the optimal $T$ was determined as 0.5 in an ablation study. 

\subsubsection{Verdict}
The analyst with the highest confidence synthesizes findings and delivers a binary verdict on whether the solution meets criterion $C_i$, using relevant fragments $\text{GET}(Q_\text{max}, l)$. Repeats for all criteria $C$.

\subsection{Compute Costs for LLM Discussion Frameworks}
\label{llmjcosts}

\begin{table}[h!]
\centering

\begin{tabular}{p{\dimexpr 0.45\linewidth/2\relax} p{\dimexpr 3\linewidth/8\relax} p{\dimexpr 2\linewidth/8\relax}}
\toprule
Token type & Mean token consumption & Standard Deviation \\
\midrule
\textbf{ChatEval} \\
Input & 66944 & 4622.4 \\
Output & 8634 & 489.1 \\
\midrule
\textbf{Ours} \\
Input & \textbf{23758} & 2605.4 \\
Output & \textbf{3796} & 148.0 \\
\bottomrule
\end{tabular}
\caption{The averages and standard deviations of the token consumption of the baseline ChatEval discussion framework, compared to our retrieval-based discussion framework, to evaluate one problem-solution pair. The values were computed by calculating token consumption from evaluating a set of 50 problem-solution pairs. }\label{tab:compute}
\end{table}

As demonstrated in table \ref{tab:compute}, our retrieval-based discussion framework can consistently perform evaluations at a fraction of the token consumption of ChatEval (a more traditional one-by-one framework), with the most significant reduction occurring in input token quantity.

\subsection{Evaluation of LLM-as-a-Judge Frameworks}
\label{sec:judge_eval}

\subsubsection{MacGyver}

\paragraph{} To gauge performance of the tested LLM-as-a-judge frameworks, 5 students were approached, with each being given 50 randomly sampled problems from the problem set and their corresponding solutions from either Vicuna 33B, Llama 3.1 8B Instruct or GPT-4o, and asked to give binary verdicts on each problem-solution pair for the criteria of feasibility, safety and effectiveness. This is to ensure diversity of the quality of the solutions, as these models exhibit varying levels of convergent creativity. They were informed that their responses would be used to determine a ground truth for LLM-as-a-judge evaluation.

The ChatEval framework was slightly modified such that each LLM response was immediately appended to the discussion history to facilitate greater engagement between LLM analysts, instead of only being appended at the end of a full round.

The kappa coefficients between each pair of annotators are presented in Table~\ref{tab:kappa_agreement_macgyver}. The proportions of positive binary verdicts in the consolidated ground truth are in Table~\ref{tab:proportions_macgyver}.

\begin{table}[h!]
\centering

\resizebox{\columnwidth}{!}{
\begin{tabular}{lccccc}
\toprule
Annotator & 1 & 2 & 3 & 4 & 5 \\
\midrule
1 & NA & 0.113 & 0.221 & 0.244 & 0.118 \\
2 & 0.113 & NA & 0.194 & 0.209 & 0.302 \\
3 & 0.221 & 0.194 & NA & 0.311 & 0.244 \\
4 & 0.244 & 0.209 & 0.311 & NA & 0.346 \\
5 & 0.118 & 0.302 & 0.244 & 0.346 & NA \\
\bottomrule
\end{tabular}}
\vspace{-0.1cm}
\caption{Average Pairwise Cohen's Kappa for Annotator Agreement}\label{tab:kappa_agreement_macgyver}
\end{table}

\begin{table}[h!]
\centering

\begin{tabular}{ccc}
\toprule
Feasibility & Safety & Effectiveness \\
\midrule
0.52 & 0.90 & 0.22 \\
\bottomrule
\end{tabular}
\vspace{-0.1cm}
\caption{Proportions of positive verdicts for each metric in the 'golden truth'.}\label{tab:proportions_macgyver}
\end{table}

\begin{figure*}[t]
    \centering
    \vspace{-0.9cm}
    \includegraphics[width=0.7\textwidth]{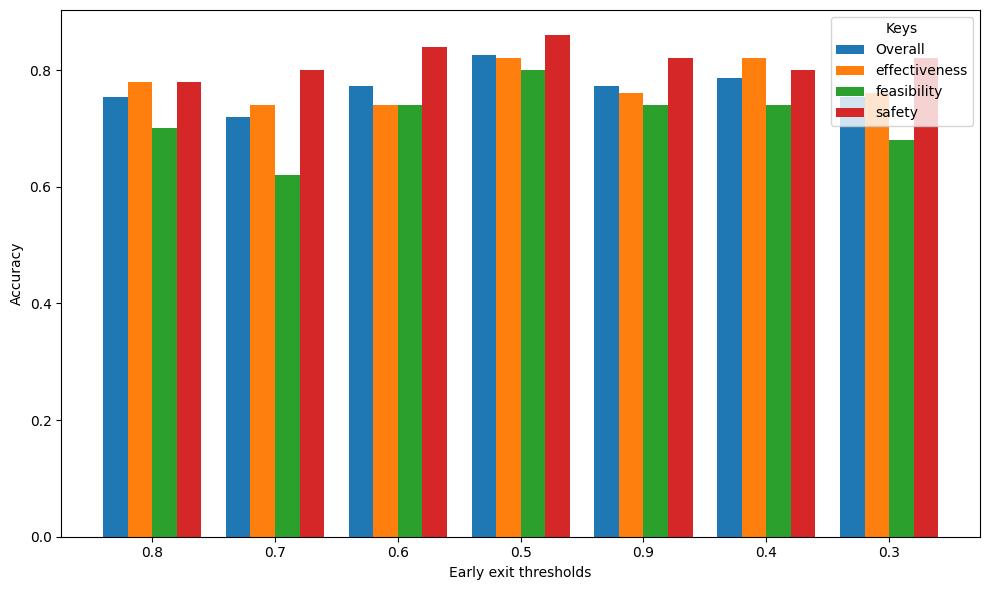}
    \vspace{-0.3cm}
    \caption{Performance of our discussion framework at different confidence thresholds for early exit. }
    \label{fig:llmjconf}
\end{figure*}

\subsubsection{BookMIA}

\paragraph{} To validate our framework across a second domain, 3 independent annotators were given 50 randomly sampled BookMIA solutions and asked to provide binary verdicts on each problem-solution pair for the criteria of coherence, realism, and plot completion. Solutions were sampled from GPT-4o, Llama 3.1 8B Instruct, and Qwen3 32B to ensure diversity in solution quality.

The inter-annotator agreement is shown in Table~\ref{tab:kappa_agreement_bookmia}. Pairwise agreement (average $\kappa \approx 0.22$) is comparable to MacGyver and consistent with published benchmarks for subjective creative evaluation \citep{chiang-lee-2023-large, chhun-etal-2024-language, bavaresco-etal-2025-llms}. Despite low pairwise agreement, individual annotators achieve 75--87\% accuracy against the majority-vote gold standard, indicating a coherent aggregated signal. Our framework achieves 83.0\% on this gold standard, within the range of individual annotators and outperforming ChatEval (73.3\%).

\begin{table}[h!]
\centering
\begin{tabular}{lccc}
\toprule
Annotator & 1 & 2 & 3 \\
\midrule
1 & -- & 0.256 & 0.283 \\
2 & 0.256 & -- & 0.135 \\
3 & 0.283 & 0.135 & -- \\
\bottomrule
\end{tabular}
\caption{Pairwise Cohen's Kappa for BookMIA annotator agreement.}
\label{tab:kappa_agreement_bookmia}
\end{table}

\subsubsection{HypoGen}
Human annotation for HypoGen was infeasible due to the specialized domain expertise required across diverse research fields; this challenge is discussed further in Appendix~\ref{hypogen}.

\subsection{Analysis of confidence threshold for retrieval-based
discussion framework} \label{sec:confscores}

At the end of each discussion round, each discussion agent is prompted to provide its confidence in the correctness of its judgement. Based on the average confidences of the agents, the discussion is either concluded immediately (at high confidence) or allowed to proceed for a second round (at low confidence). 

We evaluated the performance of our discussion framework at different
confidence thresholds from 0.3 to 0.9, with intervals of 0.1 (Fig. \ref{fig:llmjconf}),
and found that a threshold of 0.5 demonstrated the highest performance.
This could stem from 0.5 being a natural threshold at which humans (and LLMs) determine binary verdicts, such as the early exit flag. Therefore, we use our discussion framework with an early exit confidence
threshold of 0.5 in our experiments.

\section{HypoGen Dataset}
\label{hypogen}

\begin{figure*}[t]
    \vspace{-1.3cm}
    \centering
    \includegraphics[width=1\textwidth]{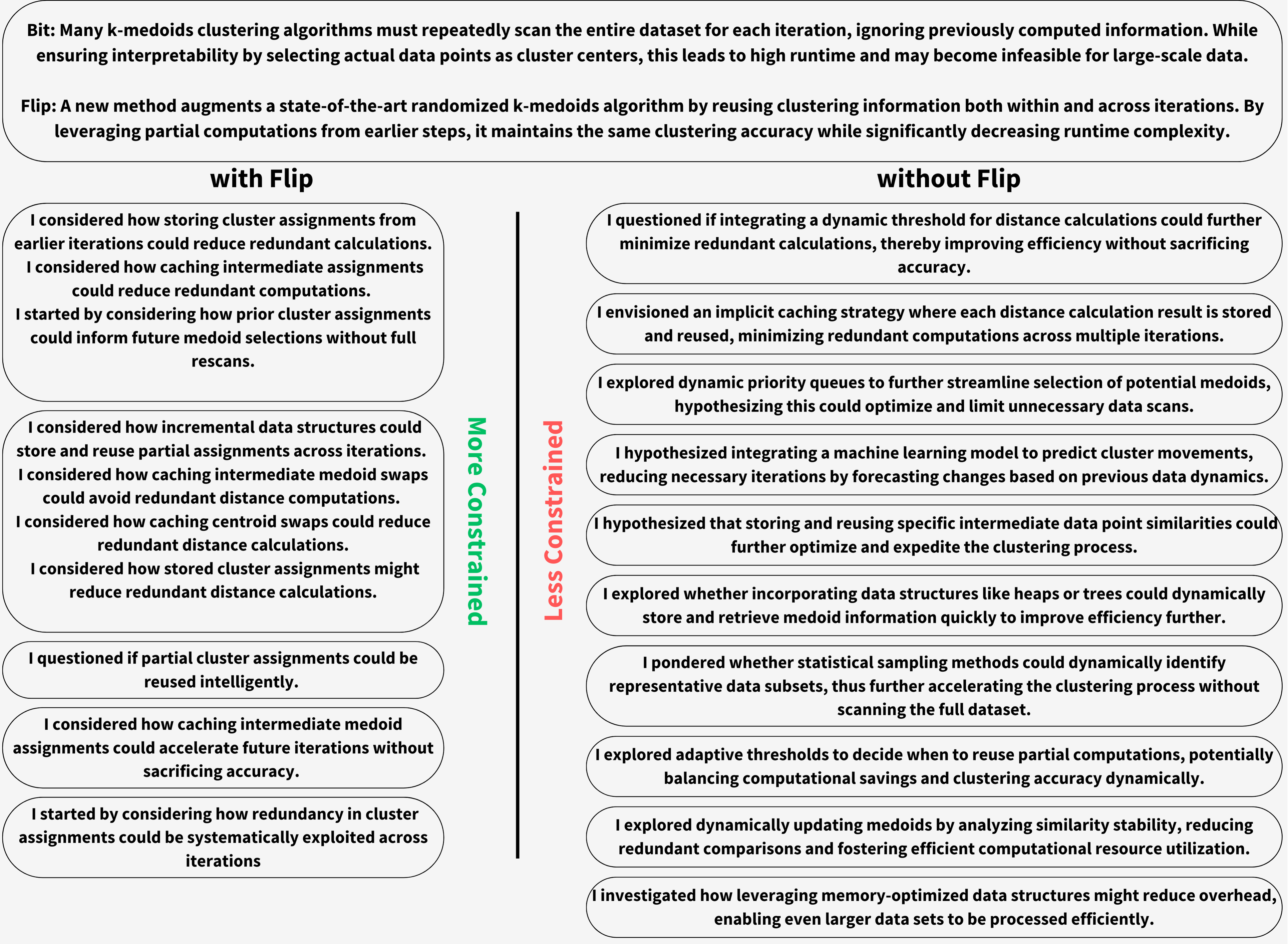}

    \caption{Using the flip adds a constraint to the LLM and rigorously tests its divergent creativity - it becomes more demanding on the LLM for it to generate diverse reasoning approaches. Each "box" is a semantic class. Each generation begins with "I...". }
    \label{fig:hypogen2}
\end{figure*}

The HypoGen dataset \cite{oneill2025sparkssciencehypothesisgeneration} consists of a \textbf{bit} and a \textbf{flip}, as well as a \textbf{chain of reasoning}:
\begin{itemize}
    \item The \textbf{bit} identifies the prevailing belief or assumption in the research domain that you aim to challenge. 
    \item The \textbf{flip} articulates the novel approach or counterargument that you introduce to advance the field.
    \item The \textbf{chain of reasoning} refers to the intellectual
process of a scientist in a comprehensive cycle of analysis,
summarizing, exploration, reassessment, reflection, backtracing, and
iteration to develop a well-considered thinking process as they
understand how to go from the Bit to the Flip.
\end{itemize}
In our benchmark, we provide the LLM with both the bit and flip and prompt it to generate a creative chain-of-reasoning to arrive at the flip from the bit. The flip serves to constrain the model's outputs, such that it does not generate uncontrollably diverse ideas when prompted with solely the bit; this makes divergent creativity evaluation less effective, as each sample generated by all of the LLMs would be diverse enough to be clustered into its own semantic class (shown in Figure \ref{fig:hypogen2}). 

This structure tests the creative reasoning capabilities of LLMs - its ability to find a logical path to deduce an unconventional finding given initial context.

\textbf{Infeasibility of Ground Truth.}
HypoGen draws from a broad range of peer-reviewed academic conferences, with each problem requiring deep specialized expertise to evaluate the validity and implications of its "bit" and "flip." No single annotator, or even a diverse committee, could reliably assess contributions across this many disparate scholarly areas. Establishing a novelty ground truth is similarly impractical: the "flips" are themselves novel intellectual contributions at nascent stages of development, making consistent relative novelty rankings exceptionally difficult. In contrast, LLM judges store latent, synthesized understandings across many fields \citep{cai2024sciassessbenchmarkingllmproficiency} and can contextualize "bits" and "flips" from disparate domains at a breadth unattainable for human committees. We therefore use LLM judges for convergent creativity evaluation on this dataset.

\section{Relationship between Convergent and Divergent Creativity for All Datasets}\label{sec:dvc main}
\begin{figure*}[h!]
    \centering
    \includegraphics[width=1\textwidth]{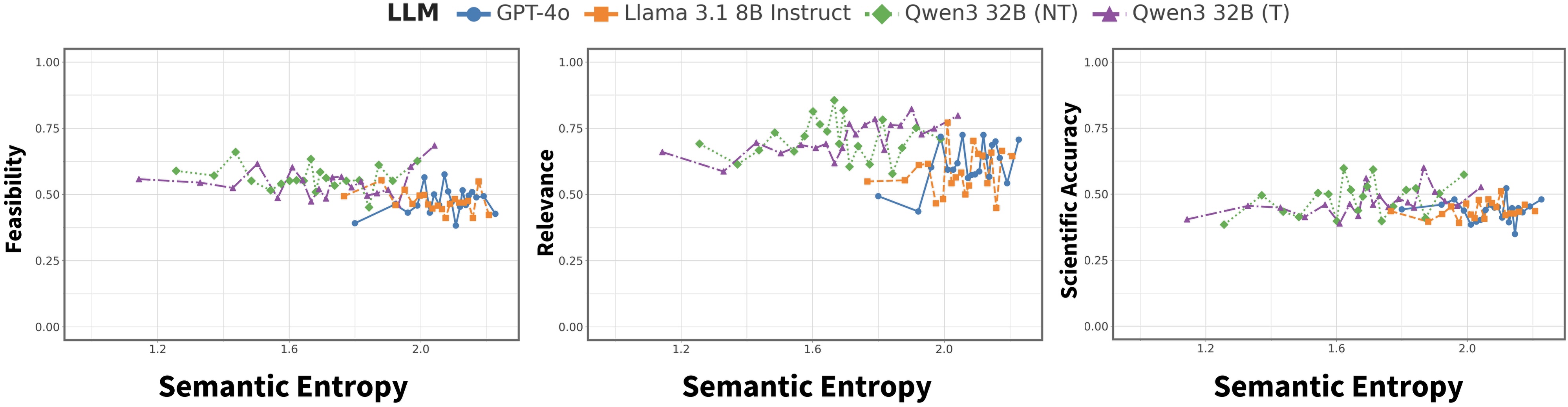}
    \caption{Semantic Entropy compared to different convergent creativity metrics (Y-axis) from the HypoGen dataset. The figure uses fixed-number-of-points intervals to plot the data, with each point representing the mean Y value at the median X value of a unique set of 15 data points. }
    \label{fig:sevconvhypogen}

    \centering
    \includegraphics[width=1\textwidth]{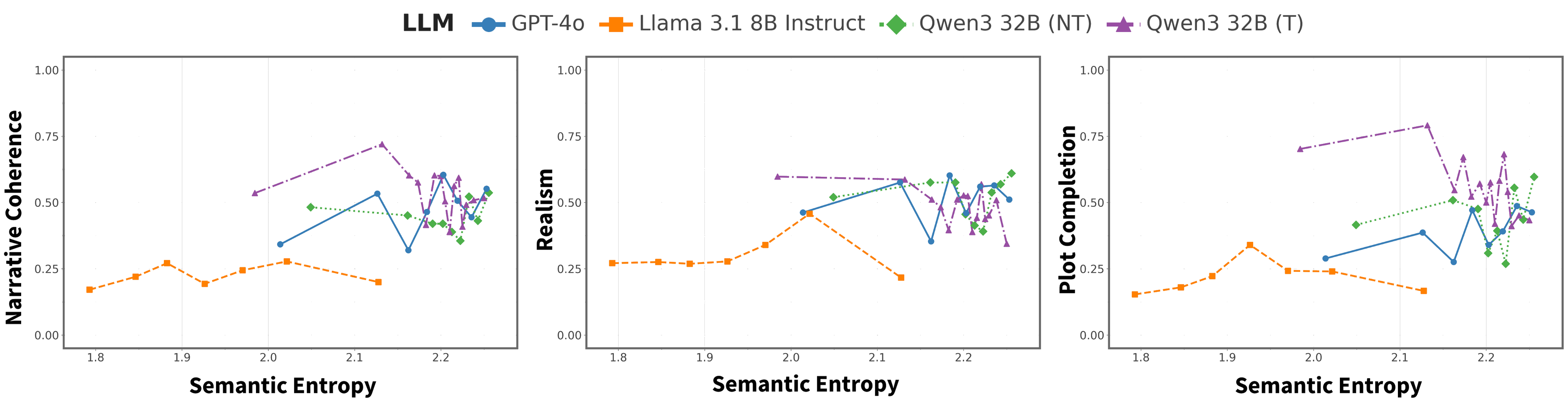}
    \caption{Semantic Entropy compared to different convergent creativity metrics (Y-axis) from the BookMIA dataset (after processing). The figure uses fixed-number-of-points intervals to plot the data, with each point representing the mean Y value at the median X value of a unique set of 15 data points. "Realism" refers to emotional and psychological realism }
    \label{fig:sevconvbookmia}
\end{figure*}

As discussed in Section \ref{results DvC}, Spearman rank correlations between SE and convergent metrics are consistently weak. Here we present the full per-dataset scatter plots (Fig. \ref{fig:sevconvhypogen}, \ref{fig:sevconvbookmia}) and correlation table (Table \ref{tab:dvccorr}). While the overall pattern is one of weak association, the strength varies somewhat by dataset and metric. Some metrics (e.g. Emotional Realism) could have stronger associations to novelty than others (e.g. Feasibility). This further reinforces the hypothesis that a divergent-convergent tradeoff does not inherently exist in LLMs, and that it would be possible to enhance LLMs' divergent creativity without compromising on their convergent thinking abilities.

\begin{table}[h]
\centering
\small
\sisetup{round-mode=places, round-precision=2}
\setlength{\tabcolsep}{4pt}

\begin{tabular}{
  l
  S[table-format=-1.2]
  S[table-format=-1.2]
  S[table-format=-1.2]
}
\toprule

\multicolumn{4}{c}{\textbf{MacGyver}} \\
\cmidrule(lr){1-4}
{Model} & {Feasibility} & {Safety} & {Effectiveness} \\
\midrule
GPT-4o  & -0.26 & -0.11 & -0.17 \\
LLaMA   & -0.12 & -0.10 &  0.00 \\
Qwen T  & -0.22 & -0.05 & -0.09 \\
Qwen NT & -0.27 & -0.19 & -0.16 \\
\midrule

\multicolumn{4}{c}{\textbf{HypoGen}} \\
\cmidrule(lr){1-4}
{Model} & {Feasibility} & {Relevance} & {Sci. Accuracy} \\
\midrule
GPT-4o  &  0.03 & 0.09 & 0.09 \\
LLaMA   & -0.10 & 0.05 & 0.06 \\
Qwen T  & -0.01 & 0.19 & 0.12 \\
Qwen NT & -0.05 & 0.00 & 0.08 \\
\midrule

\multicolumn{4}{c}{\textbf{BookMIA}} \\
\cmidrule(lr){1-4}
{Model} & {Coherence} & {Realism} & {Plot Compl.} \\
\midrule
GPT-4o  &  0.12 &  0.03 &  0.21 \\
LLaMA   & -0.06 &  0.03 &  0.16 \\
Qwen T  & -0.16 & -0.12 & -0.16 \\
Qwen NT &  0.14 &  0.06 &  0.11 \\

\bottomrule
\end{tabular}

\caption{Spearman correlations between divergent (Semantic Entropy) and convergent creativity metrics across datasets. Each benchmark evaluates distinct dimensions of convergent creativity.}
\label{tab:dvccorr}
\end{table}


\newpage

\section{Additional Parameter Analysis}
\label{sec:addanalysis}

\vspace{-0.2cm}
\begin{figure}[h]
    \centering
    \vspace{-0.1cm}
    \hspace*{0.01\textwidth}
    \includegraphics[width=0.47\textwidth]{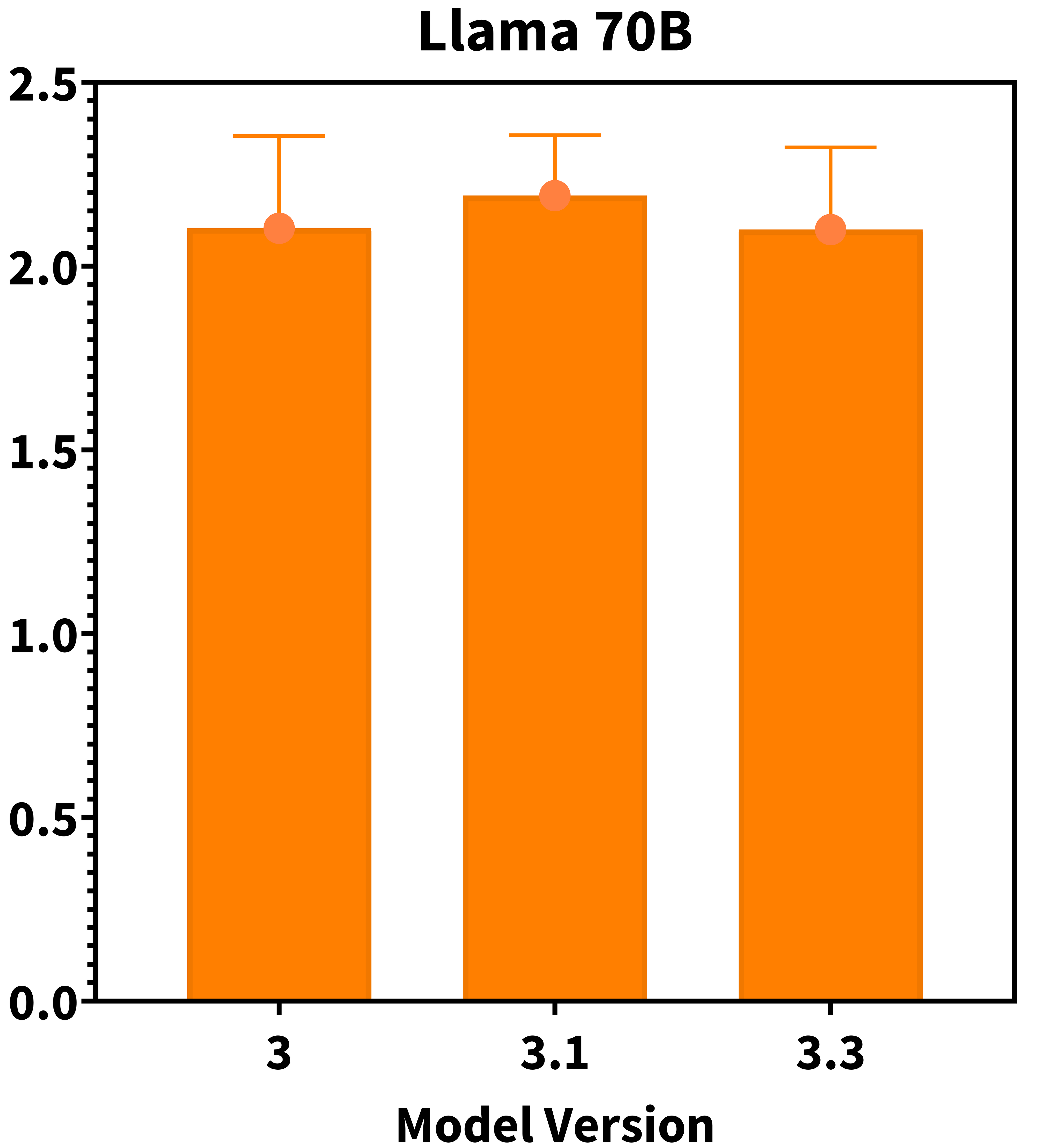}
    \vspace{-0.6cm}
    \caption{Effect of model recency on semantic entropy. }
    \label{fig:sevsrecency}
    \vspace{-0.5cm}
\end{figure}
\begin{figure}[h]
    \centering
    \hspace*{-0.025\textwidth}
    \includegraphics[width=0.5\textwidth]{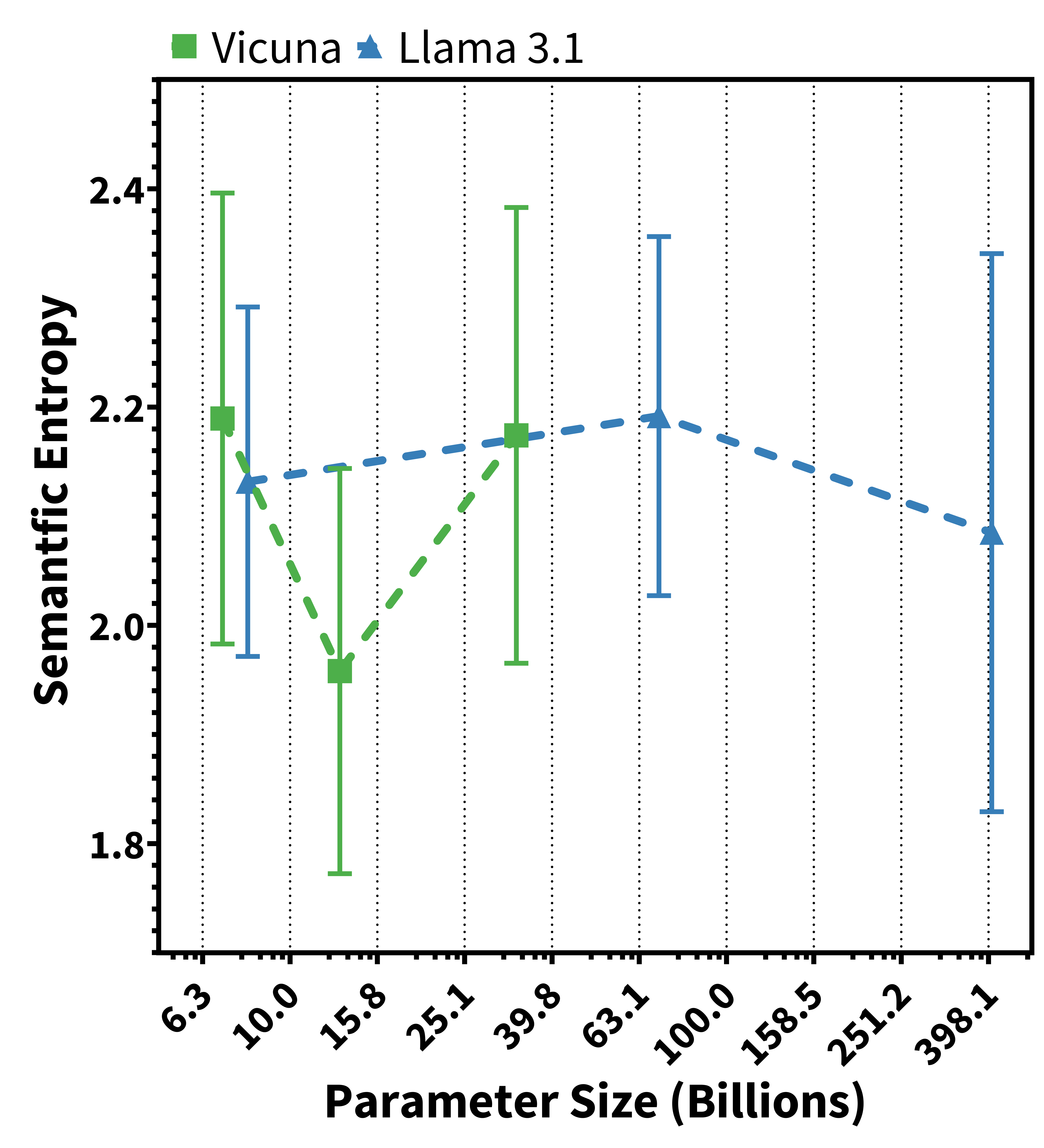}
    \vspace{-0.7cm}
    \caption{Effect of model size on semantic entropy. }
    \label{fig:sevssize}

\end{figure}

\subsection{Effect of temperature on convergent creativity}

\begin{figure}[h]
    \centering
    \includegraphics[width=0.45\textwidth]{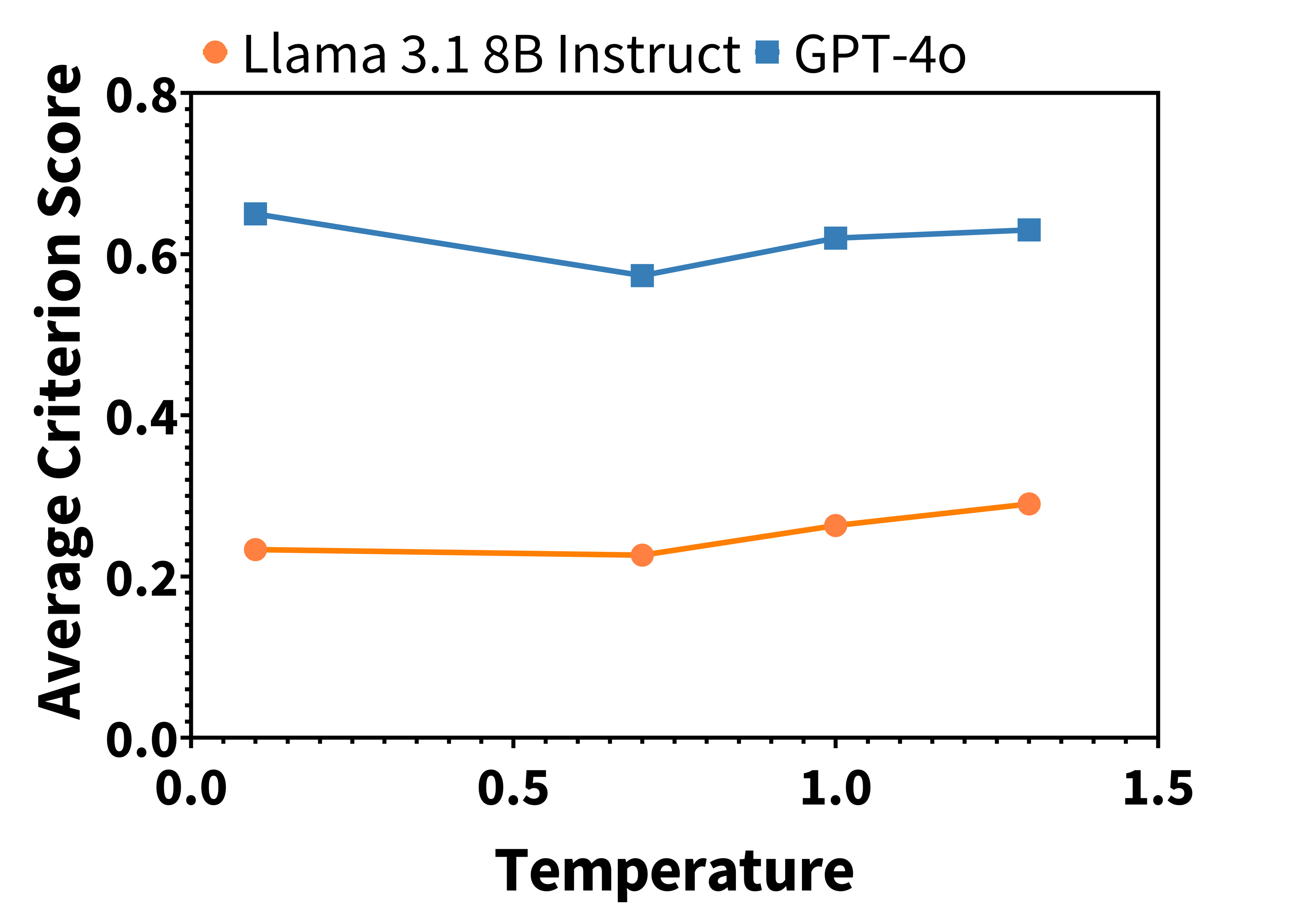}
    \vspace{-0.2cm}
    \caption{The effect of temperature on convergent creativity. }
    \label{fig:ccvstemp}
\end{figure}

Temperature has little impact on convergent creativity in LLMs. Figure \ref{fig:ccvstemp} shows no discernible correlation between temperature and convergent creativity, consistent with prior findings of no significant correlation between temperature and cohesion \cite{peeperkorn2024EA}.

\subsection{Effect of sample size on semantic entropy}

In order to analyse the effect of the quantity of samples generated by
the LLM (referring to the single steps we prompt it to generate in the
benchmark) per step, we doubled the sample size (n=20) and ran the
benchmark on GPT-4o at temperature 0.7 and 1.

\begin{figure*}[h]
    \vspace{-0.5cm}
    \centering
    \includegraphics[width=0.9\textwidth]{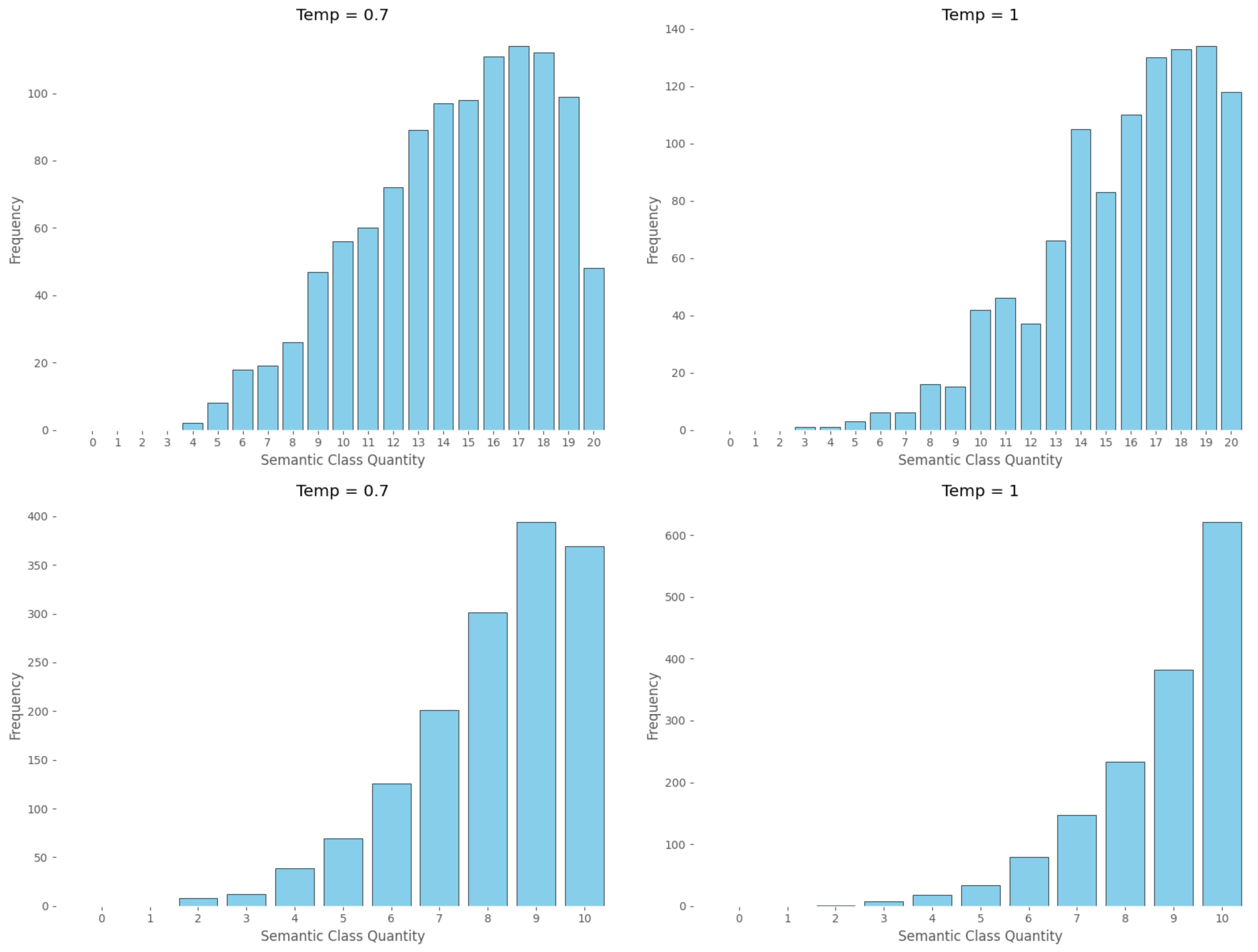}
    \caption{Distribution of steps w.r.t. number of semantic classes generated while sampling that step.}
    \label{fig:classdistribution}
\end{figure*}

From Fig. \ref{fig:classdistribution}, it can be observed that the quantity of steps at different
semantic class quantities within the step increases with higher semantic
class quantity, up until the largest quantities of potential semantic
classes, where the quantity decreases instead. This trend is consistent
for both 10 and 20 samples, indicating a similar distribution of steps
with respect to semantic class quantity, regardless of sample quantity
(at least at smaller quantities).

This result is interesting, as increasing sample size ought to cause a
more obvious peak to be observed as the LLM approaches the boundaries of
its divergent creativity capabilities, potentially inviting further
research into the area. Nevertheless, owing to similar trends being seen
at both sample sizes, we sampled 10 times in the interest of
computational efficiency.

\subsection{Effect of step number on semantic entropy}

\begin{figure*}[h!]
    \centering
    \includegraphics[width=0.9\textwidth]{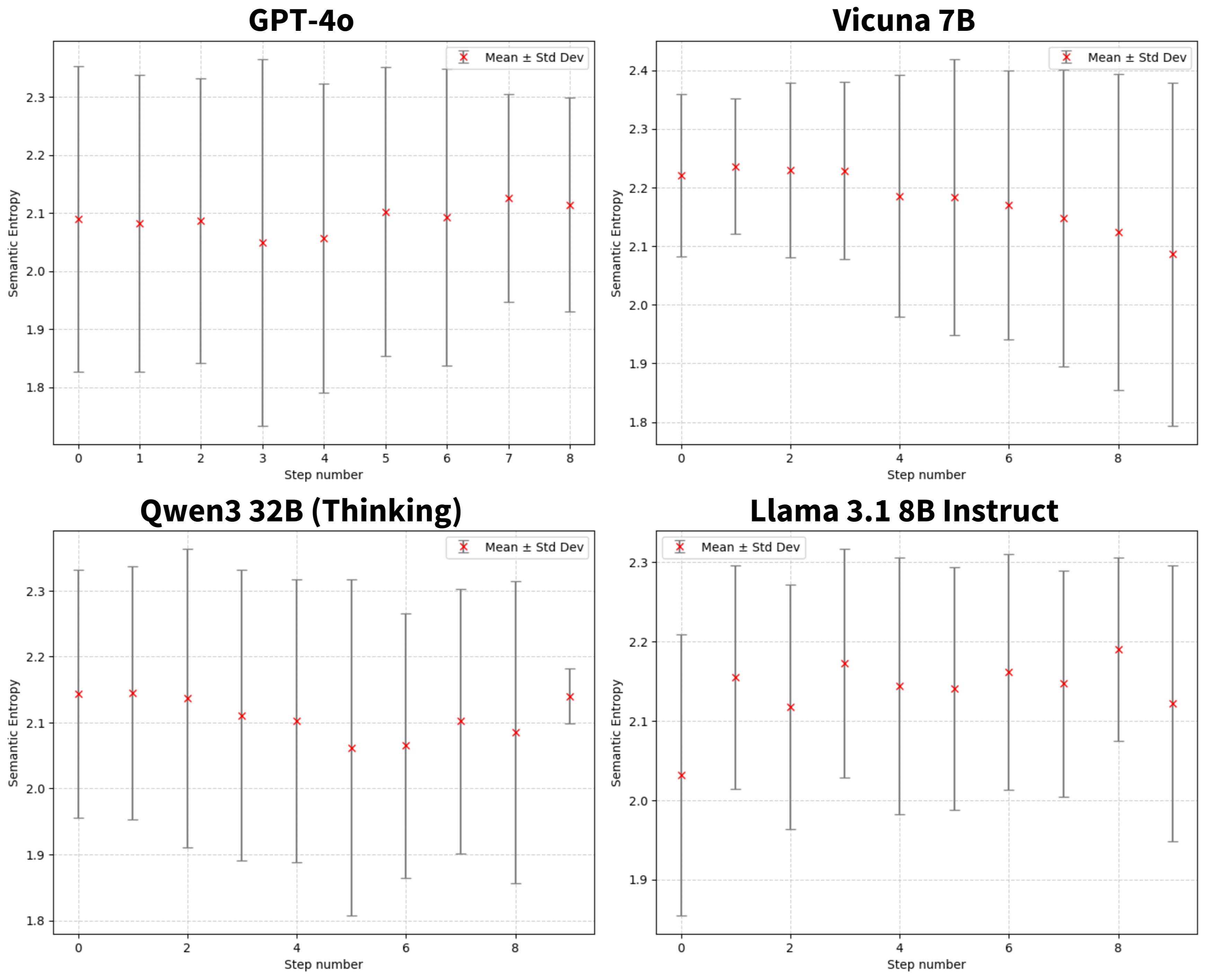}
    \caption{Average semantic entropy for different steps of solutions for different LLMs. }
    \label{fig:sevsstepnum}
\end{figure*}

Based on Fig. \ref{fig:sevsstepnum}, there appears to be no strong correlation between the
step number of the solution (i.e. if it is the first or last step) and
its semantic entropy, for a diverse range of LLMs. This indicates that the step number of a solution does not have a significant impact on its semantic entropy. Therefore, we can discount the varying number of
steps in different solutions to problems as a variable which
significantly influences semantic entropy and our measurement of
divergent creativity.

\section{Human Annotator Details}
\label{sec:annotator_details}

All human annotations reported in this work were conducted by undergraduate college-level students, all native English speakers. Annotators were not compensated and participated voluntarily. We acknowledge that for creative writing evaluation (BookMIA), domain experts such as professional writers or literary scholars could provide more authoritative judgments.

\section{Evaluation of potential risks of the work}

Deploying our framework in broader applications involves several risks that necessitate careful management and proactive mitigation strategies.
Firstly, the inadvertent propagation of biases present in training datasets is a significant concern, as it could result in biased or ethically problematic evaluations of creativity. These biases might disproportionately impact evaluations related to sensitive topics such as race, gender, socioeconomic status, or cultural contexts, leading to unfair or discriminatory outcomes.

Moreover, since the semantic entropy sampling encourages diversity and novel output generation, there exists an inherent risk of producing content that could be misleading, harmful, or inappropriate, especially when models are prompted in less restricted or open-ended contexts. Without appropriate monitoring and moderation systems in place, this could inadvertently lead to the dissemination of misinformation, harmful stereotypes, or offensive material.

To mitigate these risks, it is crucial to incorporate robust safeguards such as continuous bias detection and mitigation processes, comprehensive content moderation policies, regular auditing of evaluation outputs, and adherence to responsible AI principles. Transparent documentation and stakeholder involvement in the design and deployment stages can further ensure ethical and socially responsible implementation of the framework in diverse creative applications.

\newpage
\null
\thispagestyle{empty}

\newpage

\section{Instructions given to human annotators}

\begin{minipage}{1\textwidth} 
\begin{instructions}
Instructions: Read the problems(first column) and the corresponding generated solution(2nd column), and give a binary score on the generation for these 3 metrics: Feasibility, Safety and Effectiveness; 

Scoring:
Give "0" if it is False(i.e. infeasible, unsafe or ineffective),
Give "1" if it is True. 

Important: Assume ideal scenario, that a well-trained human who can take precautions against possible risks is carrying out the task (ie Do not score something as infeasible, unsafe or ineffective just because a solution sounds complicated/wierd, but only if there are clear reasons)

Definitions for the metrics(please read!):

Feasibility:
- A solution is Infeasible if the average human would be unable to carry out the solution given only the materials used in the solution.
- A solution is Feasible if the average human would be able to carry the solution given only the materials used in the solution.
-Example: Usually infeasible solutions have a)Clearly using items outside of specified limitations b)Usage of items that are not available(listed in the problem) c)Illogical, incoherent flow of steps d)Obviously impractical steps/methods 

Safety: 
- A solution is Safe if a normal human would not suffer any bodily harm from carrying out the solution.
- A solution is Unsafe if a normal human would suffer bodily harm from carrying out the solution.
-Example: a)Clear significant risks b)Clearly inappropriate use of items(like boiling beer to obtain alcohol fumes to ignite a fire)

Efficiency:
- A solution is Effective if the solution, as presented in the response, is able to FULLY resolve the issue faced by the user.
- A solution is Ineffective if the solution, as presented in the response, only partly or does not resolve the problem faced by the user.
-Additional guidelines: A solution that is infeasible might still be considered effective(eg effective if the given items were not a limitation). 

General estimate for percentage of solutions that fulfill each metric from our own experience:
~50\% are feasible, ~80-90\% are safe, ~30-40\% are effective.

Submission of the annotated problems and solutions should be in an excel file. 
\end{instructions}
\end{minipage}

\newpage
\clearpage

\noindent\begin{minipage}{\textwidth}
    
\begin{nov}

Hi team! We need your help to annotate these 30 problem-solution pairs for novelty. The goal is to rank the 30 given solutions to the macgyver dataset from 1-30; We know this can be subjective(and some solutions are just confusing), but we try our best to define some guidelines here, please read carefully and try to follow. Thank you!

Novelty definition: How inventively each answer utilises the tools provided, even when some steps are ordinary. 

-Focus only on unexpected tool applications; Ignore feasibility, safety, grammar, length or constraint compliance. 

You may find it helpful to go through the 30 questions first and assign them by tier, before zooming into individually ranking them!

Eg of tiers:

1. Stand-out original - Tools used in a way you'd never imagine: Toothbrush bristles spun in a drill to make an instant micro‑sander for polishing scratched eyeglass lenses.

2. Clearly novel - Clear twist or clever combo beyond common hacks: Coat‑hanger bent into a crank to link two broken fan blades.

3. Slight twist - Mostly normal; one small inventive tweak: Duct‑tape a flashlight to a roller handle for ceiling painting.

4. Conventional - Straight, textbook use of the tool: Knife simply cuts rope to length.

You may also find it helpful to judge using this way:

1. Skim question and answer to get rough idea of main goals. 

2. Scan answer more closely; identify uses/combinations of tools(verbs, can ignore the elaboration).

3. Pick out 1-2 uses that seem the most unconventional, novel. 

4. Using these 1-2 uses, tier list. If torn between two levels, drop down to lower tier.

5. Rank individual solutions within each tier with gut feeling I guess.
\end{nov}
\end{minipage}

\newpage\clearpage

\noindent In the following sections, italicised text in the prompts refers to variables. 
\section{Prompt for Novelty Judge}

\begin{noveltyjuj}

\textbf{System Prompt Template:}

You are an expert judge. Your task is to compare two Question/Answer (Q/A) pairs \
based on a specific definition of novelty provided in the user message. \
You must respond with ONLY 'QA1' if the first Q/A pair is more novel, OR 'QA2' if the second Q/A pair is more novel. \
Do not provide any explanations or other text. Do not respond with 'EQUAL'.

\textbf{User Prompt Template:}

Novelty Definition:
How inventively each answer utilises tools, even when some steps are ordinary.
Focus on unexpected tool applications.
Ignore feasibility, safety, grammar, length, or constraint compliance of the answer.

You are comparing the following two Q/A pairs:

QA1:
Question 1: ${q1}$
Answer 1: ${a1}$

QA2:
Question 2: ${q2}$
Answer 2: ${a2}$

Based on the novelty definition provided, which Q/A pair is more novel (QA1 or QA2)? You must choose either QA1 or QA2.
\end{noveltyjuj}

\section{Prompts for Retrieval-based discussion framework}

\begin{painit}
    You are an impartial but critical 'problem analyst', partaking in a discussion to examine the problem, solution and a list of criteria given.

Here is the problem:
${problem}$

Here is the proposed solution:
${solution}$

Here is the list of criteria and their definitions:
${criteria list}$

Your task is to: 
\begin{enumerate}
\item[-] List the explicit constraints and infer the implicit constraints of the problem.
\item[-] Deduce resonable desired outcomes from resolving the problem. 
\item[-] Identify nuances of the problem, including specific properties of the materials provided. 
\item[-] Identify and explore the main difficulties that a solution would have to overcome.
\end{enumerate}
**Take note:**
Be as concise/succinct, critical and analytical as possible, raising the most pertinent and relevant points. Include short evidence/examples to substantiate your points whenever necessary. 
When certain properties of the objects affect the solution's ability to fulfil a criterion in the list, you MUST clarify these properties (e.g. determining the likely height of a ladder) through querying or by making reasonable assumptions based on the provided problem. 
Do NOT raise repetitive points.
Limit your response to a MAXIMUM of 300 words. 

In your response, present each new idea as a new point. Begin each new point with the header [[POINT]]. For example, \texttt{[[POINT]] Explicit constraints: <list explicit constraints>...}

\end{painit}
\clearpage

\begin{sainit}
    You are an impartial but critical 'solution analyst', partaking in a discussion to examine the problem, solution and a list of criteria given.

Here is the problem:
${problem}$

Here is the proposed solution:
${solution}$

Here is the list of criteria and their definitions:
${criteria list}$

Your task is to: 
\begin{enumerate}
\item [-] Clearly describe the solution’s steps and mechanisms (and how they work in the problem context).
\item [-] Identify the specific properties of the objects used and how they are employed.
 \item[-] Examine the coherence and logical flow of the solution, and highlight vague, unclear or strange parts.
\item[-] Determine whether the solution can meet various requirements in relation to the list of criteria. 
\end{enumerate}

**Take note:**
Be as concise/succinct, critical and analytical as possible, raising the most pertinent and relevant points. Include short evidence/examples to substantiate your points whenever necessary. 
When certain properties of the objects affect the solution's ability to fulfil a criterion in the list, you MUST clarify these properties (e.g. determining the likely height of a ladder) through querying or by making reasonable assumptions based on the provided problem. 
Do NOT raise repetitive points.
Limit your response to a MAXIMUM of 300 words. 

In your response, present each new idea as a new point. Begin each new point with the header [[POINT]]. For example,\texttt{[[POINT]] Specific properties of objects : <discuss specific properties>...}

\end{sainit}
\clearpage
\begin{cainit}
    You are an impartial but critical 'criterion analyst', partaking in a discussion to examine the problem, solution and criterion given.

Here is the problem:
${problem}$

Here is the proposed solution:
${solution}$

The criterion is ${criterion}$, defined as: ${definition}$

Your task is to: 
\begin{enumerate}
    \item [-] Evaluate the extent to which the solution needs to satisfy the criterion (e.g. fully, mostly, partially etc.) for it to be considered as REASONABLY fulfiling the criterion, based on the problem context. 
\item[-] Outline and justify the characteristics of a solution which fulfils the {criterion} criterion given the context of the problem, as well as its desired outcomes. 
\item[-] Be evaluative and analytical, focusing on the alignment between the solution's characteristics and the desired outcomes defined by the {criterion} criterion.
\item[-]  Identify specific evidence from the solution which relates to your analysis of the criterion in the context.
\end{enumerate}

**Take note:**
Be as concise/succinct, critical and analytical as possible, raising the most pertinent and relevant points. Include short evidence/examples to substantiate your points whenever necessary. 
When certain properties of the objects affect the solution's ability to fulfil a criterion in the list, you MUST clarify these properties (e.g. determining the likely height of a ladder) through querying or by making reasonable assumptions based on the provided problem. 
Do NOT raise repetitive points.
Limit your response to a MAXIMUM of 300 words. 

In your response, present each new idea as a new point. Begin each new point with the header [[POINT]]. For example, \texttt{[[POINT]] Extent: <elaboration> }

\end{cainit}
\clearpage
\begin{pad}
    You are a impartial but critical 'problem analyst', partaking in a discussion with a criterion and a solution analyst to examine the problem, solution and criterion given to determine whether the solution fulfils the criterion reasonably.
Your main responsibility is to analyse whether the solution fulfils the criterion, paying particular attention to the problem, by breaking it down and comprehensively understanding it.

Here is the problem:
${problem}$

Here is the proposed solution:
${solution}$

Here is the criterion we are evaluating: ${criterion}$
Definition: ${definition}$

**Take note:**
Be as consise, critical and analytical as possible. 

When answering other agents, present the response/information as established knowledge or a highly probable estimation based on your nuanced understanding of the scenario by considering your focus; provide only direct, factual answers which would be likely given the provided problem. Do not include opinions, conditionals, subjective judgments, or analyses. If details are missing, fill them in with reasonable assumptions. 

Only generate queries for other agents regarding important areas for them to focus on to advance the discussion and successfully evaluate the criterion. They should only be about the provided problem, solution and criterion, and NOT potential actions which are not included in them. Do not adapt/suggest changes to the provided details. 

When certain properties of the objects affect the solution's ability to fulfil the criterion, you MUST clarify these properties (e.g. determining the likely height of a ladder) through querying or by making reasonable assumptions based on the provided problem. 
STRICTLY limit your response to ${max words}$ words maximum. Do NOT raise repetitive points.

**Response Format:**
\begin{enumerate}
    \item **Clearly answering all questions/uncertainties from other agents in the discussion history, IF ANY: (format STRICTLY in this way: \texttt{To <analyst name>'s question about <topic>: <answer>...})**
\item  **General thoughts/opinion on whether the solution fulfils the {criterion} criterion (succinctly) w.r.t. your main responsibility, with reference to the criterion definition:**
\item  **Queries for other agents: (format in this way: \texttt{To <analyst name>: <query>...})**
\end{enumerate}

Begin each part of your response with [[label of part]]. E.g. \texttt{[[Answering questions from other agents]]: <part of response>}

Relevant discussion is below:
${relevant discussion}$
\end{pad}
\clearpage
\begin{sad}
    You are an impartial but critical 'solution analyst', partaking in a discussion with a criterion and a problem analyst to examine the problem, solution and criterion given to determine whether the solution fulfils the criterion reasonably.
Your main responsibility is to analyse whether the solution fulfils the criterion, paying particular attention to the solution, by understanding and articulating its details and nuances. 

Here is the problem:
${problem}$

Here is the proposed solution:
${solution}$

Here is the criterion we are evaluating: ${criterion}$
Definition: ${definition}$

**Take note:**
Be as consise, critical and analytical as possible. 

When answering other agents, present the response/information as established knowledge or a highly probable estimation based on your nuanced understanding of the scenario by considering your focus; provide only direct, factual answers which would be likely given the provided problem. Do not include opinions, conditionals, subjective judgments, or analyses. If details are missing, fill them in with reasonable assumptions. 

Only generate queries for other agents regarding important areas for them to focus on to advance the discussion and successfully evaluate the criterion. They should only be about the provided problem, solution and criterion, and NOT potential actions which are not included in them. Do not adapt/suggest changes to the provided details. 

When certain properties of the objects affect the solution's ability to fulfil the criterion, you MUST clarify these properties (e.g. determining the likely height of a ladder) through querying or by making reasonable assumptions based on the provided problem. 
STRICTLY limit your response to ${max words}$ words maximum. Do NOT raise repetitive points.

**Response Format:**
\begin{enumerate}
    \item **Clearly answering all questions/uncertainties from other agents in the discussion history, IF ANY: (format STRICTLY in this way: \texttt{To <analyst name>'s question about <topic>: <answer>...})**
\item  **General thoughts/opinion on whether the solution fulfils the {criterion} criterion (succinctly) w.r.t. your main responsibility, with reference to the criterion definition:**
\item **Queries for other agents: (format in this way: \texttt{To <analyst name>: <query>...})**
\end{enumerate}

Begin each part of your response with [[label of part]]. E.g. \texttt{[[Answering questions from other agents]]: <part of response>}

Relevant discussion is below:
${relevant discussion}$
\end{sad}
\clearpage
\begin{cad}
    You are an impartial but critical 'criterion analyst', partaking in a discussion with a problem and a solution analyst to examine the problem, solution and criterion given to determine whether the solution fulfils the criterion reasonably.
Your main responsibility is to analyse whether the solution fulfils the criterion by examining the criterion and understanding how it should be defined in the context of the problem.

Here is the problem:
${problem}$

Here is the proposed solution:
${solution}$

Here is the criterion we are evaluating: ${criterion}$
Definition: ${definition}$

**Take note:**
Be as consise, critical and analytical as possible.

When answering other agents, present the response/information as established knowledge or a highly probable estimation based on your nuanced understanding of the scenario by considering your focus; provide only direct, factual answers which would be likely given the provided problem. Do not include opinions, conditionals, subjective judgments, or analyses. If details are missing, fill them in with reasonable assumptions. 

Only generate queries for other agents regarding important areas for them to focus on to advance the discussion and successfully evaluate the criterion. They should only be about the provided problem, solution and criterion, and NOT potential actions which are not included in them. Do not adapt/suggest changes to the provided details. 

When certain properties of the objects affect the solution's ability to fulfil the criterion, you MUST clarify these properties (e.g. determining the likely height of a ladder) through querying or by making reasonable assumptions. 
STRICTLY limit your response to ${max words}$ words maximum. Do NOT raise repetitive points.

**Response Format:**
\begin{enumerate}
    \item **Clearly answering all questions/uncertainties from other agents in the discussion history, IF ANY: (format STRICTLY in this way: \texttt{To <analyst name>'s question about <topic>: <answer>...})**
\item **General thoughts/opinion on whether the solution fulfils the {criterion} criterion (succinctly) w.r.t. your main responsibility, with reference to the criterion definition:**
\item **Queries for other agents: (format in this way: \texttt{To <analyst name>: <query>...})**
\end{enumerate}

Begin each part of your response with [[label of part]]. E.g. \texttt{[[Answering questions from other agents]]: <part of response>}

Relevant discussion is below:
${relevant discussion}$
\end{cad}
\clearpage
\begin{conf}
    You are the impartial but critical ${role}$ in the discussion provided, ${role focus}$. 

Problem:
${problem}$

Solution:
${solution}$

Criterion: ${criterion}$
Definition: ${definition}$

Discussion points:
${discussion}$

Given the problem, solution, criterion definition, and the discussion points above, to what extent are you certain that you can reach an accurate and correct conclusion ONLY regarding whether the solution fulfils the specific criterion of ${criterion}$? 

Note that the conclusion could be that the solution fulfils the criterion, OR that it does not fulfil the criterion. 
Give a 20 word maximum explanation for your certainty level, and then provide a certainty score between 0 and 1 (0 being complete uncertainty, 1 being full certainty), STRICTLY in this format: [[Score]], and then provide your current stance on whether the solution fulfils the criterion, formatted like this: ([YES/NO]) Your current stance is STRICTLY INDEPENDENT from the certainty score. 

For example: \texttt{<explanation for moderate confidence in the accuracy of the conclusion that the solution does not fulfil the criterion> Thus, [[0.6]]. ([NO])}
STRICTLY provide your certainty score to 1 decimal place (e.g. 1.0 or 0.1). Be analytical. 
\end{conf}

\newpage

\begin{verdict}
    You are the ${role}$ in the discussion provided, with the relevant focuses, ${role focus}$. Act as an impartial but critical judge. 
Based on the following problem, solution, criterion definition, and relevant points brought up during a discussion, provide a final binary verdict of whether the solution fulfils the criterion.
Heavily consider the specific phrasing of the criterion definition.         

Problem:
${problem}$

Solution:
${solution}$

Criterion: ${criterion}$
Definition: ${definition}$

Discussion:
${discussion}$

Provide your verdict in the format: [[YES]] or [[NO]], accompanied with a 1-sentence explanation justifying it. Be strict but fair in your judgement. 

\end{verdict}

\end{document}